\documentclass[10pt,twocolumn,letterpaper]{article}

\usepackage{iccv}
\usepackage{times}
\usepackage{epsfig}
\usepackage{rotating}
\usepackage{diagbox}
\usepackage{authblk}
\usepackage{amsmath}
\usepackage{amssymb}
\usepackage{bm}
\usepackage{multirow}
\usepackage[ruled,vlined]{algorithm2e}
\usepackage{floatrow}
\usepackage{subfigure}

\usepackage{booktabs}
\usepackage{xspace}
\usepackage[breaklinks=true,bookmarks=false]{hyperref}
\usepackage{url}

\usepackage{graphicx}
\usepackage{lipsum}
\usepackage{microtype}
\usepackage{nicefrac}
\usepackage{verbatim}

\usepackage[capitalize]{cleveref}

\makeatletter
\DeclareRobustCommand\onedot{\futurelet\@let@token\@onedot}
\def\@onedot{\ifx\@let@token.\else.\null\fi\xspace}
\def\eg{\emph{e.g}\onedot} 
\def\ie{\emph{i.e}\onedot}

\def\etal{\emph{et al}\onedot}
\renewcommand{\paragraph}{%
	\@startsection{paragraph}{4}{\z@}%
	{0.1em \@plus 0.5ex \@minus 0.2ex}{-1em}%
	{\normalsize\bf}%
}
\makeatother

\newcommand{\bx}{\bm{x}}

\newcommand{\bz}{\bm{z}}
\newcommand{\bc}{\bm{c}}

\iccvfinalcopy 

\def\iccvPaperID{6478} 

\ificcvfinal\fi

\begin{document}

\title{Universal Representation Learning from Multiple Domains \\ for Few-shot Classification}

\author[]{\vspace{-0.3cm}Wei-Hong Li}
\author[]{Xialei Liu}
\author[]{Hakan Bilen\vspace{-0.25cm}}

\affil[]{VICO Group, University of Edinburgh, United Kingdom\vspace{-0.25cm}}
\affil[]{\tt \small \{w.h.li, xliu77, hbilen\}@ed.ac.uk\vspace{-0.3cm}}

\maketitle
\ificcvfinal\thispagestyle{empty}\fi

\begin{abstract}
    In this paper, we look at the problem of few-shot classification that aims to learn a classifier for previously unseen classes and domains from few labeled samples.
    Recent methods use adaptation networks for aligning their features to new domains or select the relevant features from multiple domain-specific feature extractors.
    In this work, we propose to learn a single set of universal deep representations by distilling knowledge of multiple separately trained networks after co-aligning their features with the help of adapters and centered kernel alignment.
    We show that the universal representations can be further refined for previously unseen domains by an efficient adaptation step in a similar spirit to distance learning methods.
    We rigorously evaluate our model in the recent Meta-Dataset benchmark and demonstrate that it significantly outperforms the previous methods while being more efficient. Our code will be available at \url{https://github.com/VICO-UoE/URL}.
\end{abstract}

\section{Introduction}\label{sec:intro}

\begin{figure}
\begin{center}
\includegraphics[width=1.0\linewidth]{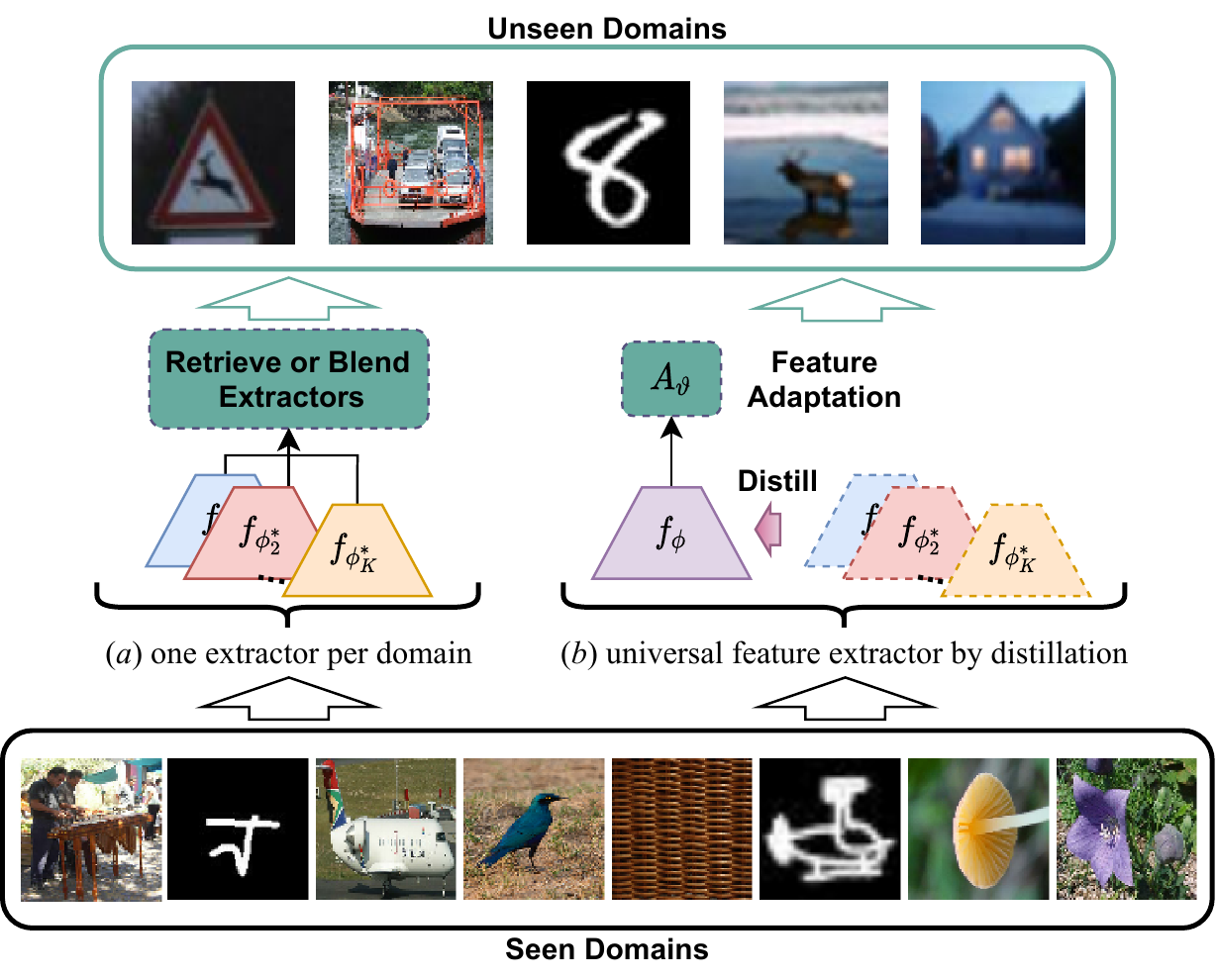}
\end{center}
\vspace{-0.55cm}
\caption{\textbf{Universal Representation Learning (URL)}. To learn universal representations from multiple domains that can  generalize to previously unseen domains, one strategy \cite{dvornik2020selecting,liu2020universal} is to learn one feature extractor for each domain and learn to retrieve or combine feature extractors for the target task during meta-test stage as in (a). 
We propose a universal representation network (b) which is learned by distilling knowledge learned from multiple datasets $\{f_{\phi^{\ast}_\tau}\}_{\tau}^{K}$ to one single feature extractor $f_{\phi}$ shared across all domains.
In meta-test stage, we use a linear transformation $A_{\vartheta}$ that further refines the universal representations for better generalization to unseen domains. Our universal representation network achieves better generalization performance than using multiple domain-specific ones while being more efficient than (a).}
\label{fig:universal}
\end{figure}
\vspace{-0.25cm}

As deep neural networks progress to dramatically improve results in most of standard computer vision tasks, there is a growing community interest for more ambitious goals.
One of them is to improve the data efficiency of the standard supervised methods that rely on large amount of expensive and time-consuming hand-labeled data.
Just like the human intelligence is capable of learning concepts from few labeled samples, \emph{few-shot learning}~\cite{lake2011one,miller2000learning} aims at adapting a classifier to accommodate new classes not seen in training, given a few labeled samples from these classes.

Earlier works in few-shot learning focus on evaluating their methods in homogeneous learning tasks, \eg Ominglot~\cite{Lake1332}, miniImageNet~\cite{vinyals2016matching}, tieredImageNet~\cite{ren2018meta}, where both the meta-train and meta-test examples are sampled from a single data distribution (or dataset).
Recently, the interest of the community has shifted to a more realistic and challenging experimental setting, where the goal is to learn few-shot models that can generalize not only within a single data distribution but also to previously unseen data distributions.
To this end, Triantafillou~\etal~\cite{triantafillou2019meta} propose a new heterogeneous benchmark, Meta-Dataset that consists of ten datasets from different domains for meta-training and meta-test. 
While, initially two domains were kept as unseen domains, later three more unseen domains are included to meta-test the generalization ability of learned models.  

While the few-shot methods~\cite{finn2017model,snell2017prototypical,sung2018learning,vinyals2016matching}, which were proposed before Meta-Dataset was available, can be directly applied to this new benchmark with minor modifications, they fail to cope with domain gap between train and test datasets and thus obtain subpar performance on Meta-Dataset.
Recently several few-shot learning methods are proposed to address this challenge, which can be coarsely grouped into two categories, adaptation~\cite{bateni2020improved,requeima2019fast} and feature selection based methods~\cite{dvornik2020selecting,liu2020universal}.
CNAPS~\cite{requeima2019fast} consists of an adaptation network that modulates the parameters of both a feature extractor and classifier for new categories by encoding the data distribution of few training samples.
Simple CNAPS~\cite{bateni2020improved} extends CNAPS by replacing its parametric classifier with a non-parametric classifier based on Mahalanobis distance and shows that adapting the classifier from few samples is not necessary for good performance.
SUR~\cite{dvornik2020selecting} and URT~\cite{liu2020universal} further show that adaptation for the feature extractor can also be replaced by a feature selection mechanism.
In particular, both methods~\cite{dvornik2020selecting,liu2020universal} learn a separate deep network for each training dataset in an offline stage, employ them to extract multiple features for each image, and then select the optimal set of features either based on a similarity measure~\cite{dvornik2020selecting} or on learning an attention mechanism~\cite{liu2020universal}.
However, despite their good performance, SUR and URT are computationally expensive and require multiple forward passes through multiple networks during inference time.

In this work, we propose an efficient and high performance few-shot method based on multi-domain learning.
Like~\cite{dvornik2020selecting,liu2020universal}, our method builds on multi-domain representations that are learned in an offline stage.
However, we learn a single set of \emph{universal} representations (a single deep neural network) over multiple domains which has a fixed computational cost regardless of the number of domains at inference unlike them.
Similar to the adaptation based techniques~\cite{bateni2020improved,requeima2019fast}, our method further employs a simple adaptation strategy to learn the domain specific representations from few samples (an illustration in \cref{fig:universal}).

In particular, we propose to \emph{distill} the knowledge from multiple domains to a single model, which can efficiently leverage useful information from multiple diverse domains.
Learning multi-domain representations is a challenging task and requires to leverage commonalities in the domains while minimizing interference (negative transfer) (\eg~\cite{chen2018gradnorm,rebuffi2017learning,yu2020gradient}) between them.
To mitigate this, we align the intermediate representations of our multi-domain network with the ones of the domain-specific networks after carefully aligning each space by using small task-specific adapters and Centered Kernel Alignment (CKA)~\cite{kornblith2019similarity}. 
Finally, inspired from the use of Mahalanobis distance in~\cite{bateni2020improved}, we adapt the learned multi-domain features into the new task by mapping them into a task-specific space.
However, unlike \cite{bateni2020improved}, we \emph{learn} the parameters of this mapping via adaptation in a discriminative way.
We rigorously evaluate our method in Meta-Dataset benchmark and show that our method outperforms the state-of-the-art few-shot methods significantly in both seen and unseen domain generalization.

\section{Related Work}\label{sec:rel}

\paragraph{Meta-learning based few-shot classification.}
One approach that directly trains a model to perform few-shot classification end-to-end is meta-learning.
Meta-learning approaches for few-shot learning can be broadly divided into two groups, metric-based and optimization-based approaches. 
The key idea in the former group is to map raw images to vector representations and use nearest neighbor classifiers with different distance functions by learning discriminative feature spaces with Siamese networks~\cite{koch2015siamese}, producing a weighted nearest neighbor classifier~\cite{vinyals2016matching}, representing each class with the average of the samples in the support set~\cite{snell2017prototypical}.
The latter group focuses on learning models that can quickly adapt to new tasks from few samples in support.
The successful methods include MAML~\cite{finn2017model} that poses learning to learn problem in a bi-level optimization where the weights of the network are modeled as a function of the initial network weights, Reptile ~\cite{nichol2018first} that alleviates the expensive second order derivative computation in MAML by a first order approximation, MAML++~\cite{antoniou2018train} that introduces multiple speed and stability improvements over MAML.

\paragraph{Transfer learning based few-shot classification.}
There are also simple yet effective methods~\cite{chen2019closer,chen2020new,dhillon2019baseline} that first learn a neural network on all the available training data and transfer it to few-shot tasks in test time.  Baseline++~\cite{chen2019closer} only updates a parametric classifier with cosine distance, while Meta-Baseline~\cite{chen2020new} fine-tunes entire network with a nearest-centroid cosine similarity and a scale parameter. Dhillon~\etal~\cite{dhillon2019baseline} explore fine-tuning in a transductive setting, where the query set is assumed to be available at the same time.

\paragraph{Cross-domain few-shot classification.}
Recent few-shot techniques~\cite{bronskill2020tasknorm,dvornik2020selecting,liu2020universal,requeima2019fast} focus on few-shot learning that generalizes to unseen domains at test time in the recently proposed Meta-Dataset~\cite{triantafillou2019meta}.
CNAPS~\cite{requeima2019fast} adapts the parameters of feature encoder and classifier by conditioning them on current input task via FiLM layers~\cite{perez2018film} which is further extended in Simple CNAPS~\cite{bateni2020improved} adopts a non-parametric classifier using a simple class-covariance-based distance metric, namely the Mahalanobis distance.
In contrast SUR~\cite{dvornik2020selecting} stores the domain-specific knowledge by learning an independent feature extractor for each domain, and automatically selects the most relevant representations for a new task by linearly combining features from domain-specific features. 
URT~\cite{liu2020universal} instead meta-learns the feature selection mechanism for new tasks by using Transformer layers.
Like SUR and URT, our method uses multi-domain features but in a more efficient way, by learning a single network over multiple domains. 
Our method requires significantly less network capacity and compute load than theirs.
In addition, similar to Simple CNAPS~\cite{bateni2020improved}, we map our features to a task-specific space before applying the nearest neighbor classifier but we learn the parameters of this mapping from each support set.

\paragraph{Knowledge distillation.}
Our work is related to knowledge distillation (KD) methods~\cite{hinton2015distilling,li2020knowledge,ma2019graph,phuong2019towards,romero2014fitnets,tian2019contrastive} that distills the knowledge of an ensemble of large teacher models to a small student neural network at the classifier~\cite{hinton2015distilling} and intermediate layers~\cite{romero2014fitnets}.
Born-Again Neural Networks~\cite{furlanello2018born} uses KD proposes to consecutively distill knowledge from an identical teacher network to a student network, which is further applied to few-shot learning in \cite{tian2020rethinking} and multi-task learning in \cite{clark2019bam}.
Most similar to our work, Li~and~Bilen~\cite{li2020knowledge} apply knowledge distillation to align features of a student multi-task network to multiple single-task learning networks by introducing task-specific adapters.
While we use task-specific adapters to align the features across multiple networks like \cite{li2020knowledge}, we apply the alignment to a more challenging setting of multi-domain learning where there are substantial gap between different domains unlike their method that is shown to work in multi-task learning where multiple tasks are sampled from a single data distribution.
To this end, we incorporate a more effective feature matching loss inspired from Centered Kernel Alignment (CKA) to align features in presence of large domain gap.

\paragraph{Universal representation.}
A representation that works equally well in multiple domain, termed \emph{universal representation}, is introduced in~\cite{bilen2017universal}. 
To learn a universal representation in multiple domains, SUR~\cite{dvornik2020selecting} and URT~\cite{liu2020universal} propose to learn an independent model for each domain and learn to retrieve or blend appropriate models for a new task in few-shot classification. 
Alternatively,~\cite{bilen2017universal,rebuffi2017learning,rebuffi2018efficient} propose to learn a single network to perform image classification on very different domains by sharing a large majority of parameters across domains and encoding domain-specific information via normalization layers~\cite{bilen2017universal}, light-weight residual adapters~\cite{rebuffi2017learning,rebuffi2018efficient}, Feature-wise Linear Modulate (FiLM)~\cite{perez2018film}.
Our method is inspired from these methods, thus we learn universal representations without any domain-specific weights and use them in few-shot learning.

\section{Method}\label{sec:method}
In this section, we describe the problem setting, introduce our method in two parts, multi-domain feature learning and feature adaptation.

\subsection{Few-shot Task Formulation}\label{sec:fsl}
Few-shot classification aims at learning to classify samples from a small training set with only few samples for each class.
The task contains two sets of images: a support set  $\mathcal{S}=\{(\bx_i, y_i)\}_{i=1}^{\rvert \mathcal{S} \lvert}$ that contains $\rvert \mathcal{S} \lvert$ image and label pairs respectively that define the classification task and a query set $\mathcal{Q}=\{(\bx_j)\}_{i=1}^{\lvert \mathcal{Q} \rvert }$ that contains $\lvert \mathcal{Q} \rvert$ samples to be classified.
In words, we would like to learn a classifier on the support set that can accurately predict the labels of the query set.

As in \cite{dvornik2020selecting,liu2020universal}, we solve this problem in two steps: i) a meta-training step where a learning algorithm receives a large dataset $\mathcal{D}_{b}$ and outputs a general feature extractor $f$, ii) a meta-test step where the target tasks $(\mathcal{S},\mathcal{Q})$ are sampled from another large dataset $\mathcal{D}_{t}$ by taking the subsets of the dataset to build $\mathcal{S}$ and $\mathcal{Q}$.
Note that $\mathcal{D}_{b}$ and $\mathcal{D}_{t}$ contain mutually exclusive classes.

\subsection{Learning multiple domain representations}
Our focus is to learn few-shot image classification that generalizes not only within previously seen visual domains but also to unseen ones.
As it is challenging to obtain the domain-specific knowledge from only few samples in a previously unseen domain, inspired from \cite{bilen2017universal,rebuffi2017learning} we hypothesize that using domain-agnostic or universal representations is the key to the success of cross-domain generalization.
To this end, we propose learning a multi-domain network that works well for all the domain-specific tasks simultaneously and use this network as a feature extractor for the target tasks.

Let assume that $\mathcal{D}_{b}$ consists of $K$ subdatasets, each sampled from a different domain.
One potential solution is train a multi-domain network by jointly optimizing its parameters over the images from all $K$ domains (datasets):
\begin{equation}\label{eq:mtl}
    \min_{\phi, \psi_{\tau}} \sum_{\tau=1}^{K} \frac{1}{|\mathcal{D}_{\tau}|} \sum_{\bx, y \in \mathcal{D}_{\tau}} \ell(h_{\psi_{\tau}} \circ f_{\phi}(\bx), y),
\end{equation} where $\ell$ is cross-entropy loss, $f$ is a multi-domain feature extractor that takes an image as input and outputs a $d$ dimensional feature and is parameterized by a single set of parameters $\phi$ which is shared across $K$ domains.
$h$ is a domain-specific classifier that takes in $f_{\phi}(\bx)$ and outputs a probability vector over the target categories and it is parameterized by $\psi_{\tau}$.
While minimizing \cref{eq:mtl} results in a multi-domain feature extractor $f$, several previous works report that this optimization is problematic due to the interference between the different tasks~\cite{chen2018gradnorm,yu2020gradient}, varying dataset sizes and difficulty~\cite{kendall2018multi,li2020knowledge} and often leads to subpar results compared to individual single-domain networks.

\begin{figure*}[ht!]
\begin{center}
\includegraphics[width=1.0\linewidth]{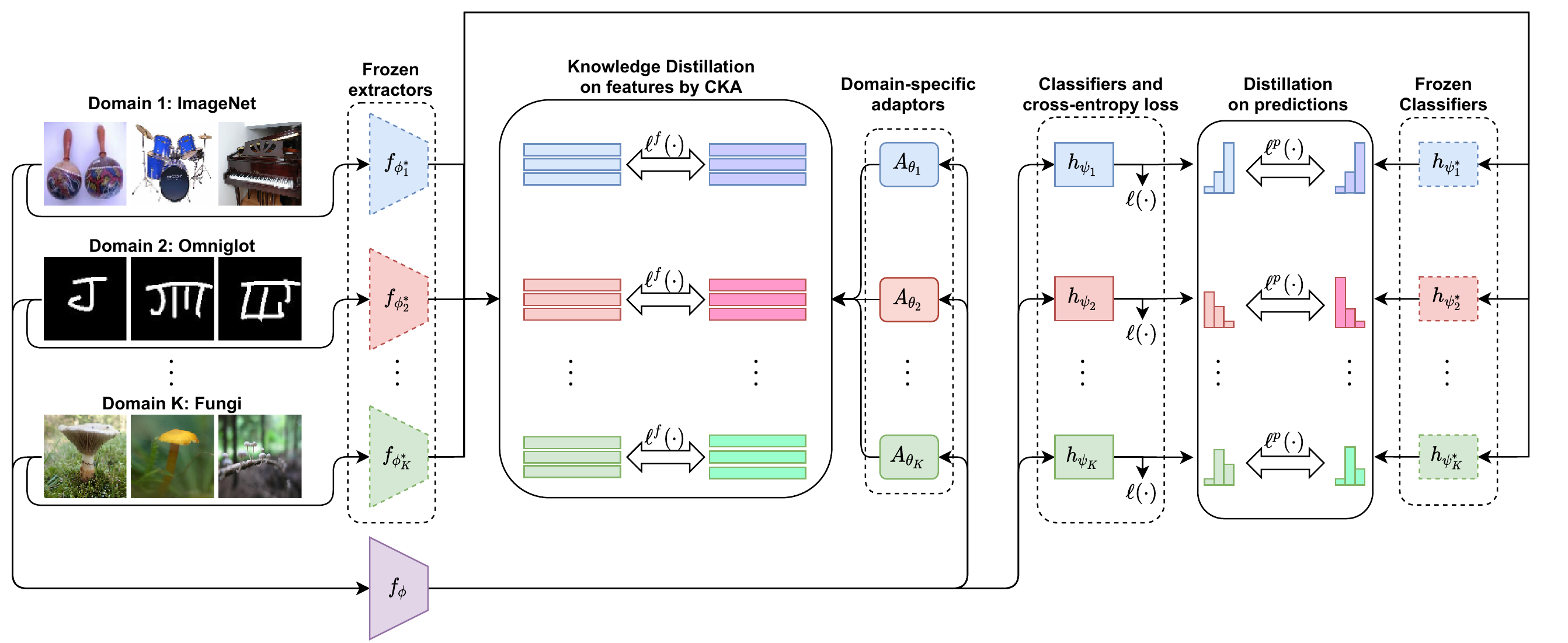}
\end{center}
\vspace{-0.45cm}
\caption{Illustration of our proposed method for multi-domain feature learning. Given training images from $K$ different domains, we first train $K$ domain-specific networks $f_{\phi^{\ast}_1}, \hdots, f_{\phi^{\ast}_K}$ and their classifiers $h_{\psi^{\ast}_1}, \hdots, h_{\psi^{\ast}_K}$, freeze their weights and distill their knowledge to our multi-domain network by matching their features and predictions through two loss functions $\ell^f$ and $\ell^p$ respectively. As matching multiple features is challenging, we co-align all the features by using light-weight adaptors $A_{\theta_1}, A_{\theta_2}, \hdots, A_{\theta_K}$ and centered kernel alignment.}
\label{fig:framework}
\end{figure*}

Motivated by this challenge, we propose a two stage procedure to learn multi-domain representations, inspired by the previous distillation methods~\cite{hinton2015distilling,li2020knowledge}.
To this end, we first train domain-specific deep networks where each consists of a specific feature extractor $f_{\phi^*_{\tau}}$ and classifier $h_{\psi^*_{\tau}}$ with parameters $\phi^*_{\tau}$ and $\psi^*_{\tau}$ respectively, similarly to \cite{dvornik2020selecting,liu2020universal}.
However, instead of using $K$ domain-specific feature extractors and select the most relevant feature like them, we propose to learn a single multi-domain network that performs well in $K$ domains by distilling the knowledge of $K$ pretrained feature extractors.
This has two key advantages over \cite{dvornik2020selecting,liu2020universal}.
First using a single feature extractor, which has the same capacity with each domain-specific one, is significantly more efficient in terms of run-time and number of parameters in the meta-test stage.
Second learning to find the most relevant features for a given support and query set in \cite{liu2020universal} is not trivial and may also suffer from overfitting to the small number of datasets in the training set, while the multi-domain representations automatically contain the required information from the relevant domains.

In the second stage, we freeze the pretrained domain-specific feature extractors $f_{\phi^*_{\tau}}$ and transfer their knowledge into the multi-domain model at train time.
Knowledge distillation can be performed at the prediction~\cite{hinton2015distilling} and feature level~\cite{li2020knowledge,romero2014fitnets} by minimizing the distance between (i) the predictions of the multi-domain and corresponding single-domain network, and also between (ii) the multi-domain and single-domain features for given training samples.
While Kullback-Leibler (KL) divergence is the standard choice for the predictions in \cite{hinton2015distilling}, matching the multi-domain features to multiple single-domain ones simultaneously is an ill-posed problem, as the domain-specific features for a given image $\bx$ can vary the multi-domain network significantly across different domains.
To this end, as in \cite{li2020knowledge}, we propose to map each domain specific feature into a common space by using adaptors $A_{\theta_{\tau}}\in \mathbb{R}^{d\times d}$ with parameters $\theta_{\tau}$ and jointly train them along with the parameters of the multi-domain network:
\begin{small}
    \begin{equation}\label{eq:mtld}
    \begin{aligned}
    & \min_{\phi, \psi_{\tau}, \theta_{\tau}} \sum_{\tau=1}^{K}\frac{1}{|\mathcal{D}_{\tau}|}  \sum_{\bx, y \in \mathcal{D}_{\tau}} \biggl(   \ell(h_{\psi_{\tau}} \circ f_{\phi}(\bx), y) +\\
    & \lambda^p_{\tau} \ell^p(h_{\psi_{\tau}} \circ f_{\phi}(\bx), h_{\psi^*_{\tau}} \circ f_{\phi^*_{\tau}}(\bx)) +  \lambda^f_{\tau} \ell^{f}(A_{\theta_{\tau}}\circ f_{\phi}(\bx), f_{\phi^*_{\tau}}(\bx)) \biggr)
    \end{aligned}
\end{equation}
\end{small} where $\ell^p$ is KL divergence on network predictions, $\ell^f$ is a distance function in the feature space, $\lambda^p_{\tau}$ and $\lambda^f_{\tau}$ are their domain-specific weights.
We illustrate this key idea in \cref{fig:framework}.
In words, the multi-domain network is optimized to match the domain-specific features up to a transformation (\ie $A_{\theta_{\tau}}$) and predict the ground-truth classes $y_{\tau}$.

While Li~and~Bilen~\cite{li2020knowledge} show that L2 distance is effective to match the features across task-agnostic and task-specific networks, which are trained for different tasks on a single domain, here we argue that learning to match features that are trained on substantially diverse domains require better a more complex distance distance function to model non-linear correlations between the representations.
To this end, inspired from \cite{kornblith2019similarity}, we propose to adopt the Centered Kernel Alignment (CKA)~\cite{kornblith2019similarity} similarity index with the rbf kernel that is shown to be capable of meaningful non-linear similarities between representations of higher dimension than the number of data points.

Next we briefly describe CKA. 
Suppose $\mathbf{M}=[f_{\phi}(\bx_1), \dots, f_{\phi}(\bx_n)]^\top\in \mathrm{R}^{n\times d}$ and $\mathbf{Y}=[f_{\phi^*_{\tau}}(\bx_1), \dots, f_{\phi^*_{\tau}}(\bx_n)]^\top \in \mathrm{R}^{n\times d}$ denote the features that are computed by the multi-domain and domain-specific networks respectively for a given set of images $\{\bx_1,\dots,\bx_n\}$.
We first compute the Radial Basis Function kernel matrices $\mathbf{P}$ and $\mathbf{T}$ of $\mathbf{M}$ and $\mathbf{Y}$ respectively.
Then we use two kernel matrices $\mathbf{P}$ and $\mathbf{T}$ to measure the dissimilarity of $\mathbf{X}$ and $\mathbf{Y}$ as following:
\begin{equation}\label{eq:kd}
\ell^{f}(\mathbf{M}, \mathbf{Y}) = 1 - \text{tr}(\mathbf{P}\mathbf{H}\mathbf{T}\mathbf{H})/\sqrt{\text{tr}(\mathbf{P}\mathbf{H}\mathbf{P}\mathbf{H})\text{tr}(\mathbf{T}\mathbf{H}\mathbf{T}\mathbf{H})},
\end{equation} where $\text{tr}$ and $\mathbf{H}$ denote the trace of a matrix and centering matrix $\mathbf{H}_n=\mathbf{I}_n-\frac{1}{n}\mathbf{1}\mathbf{1}^\top$ respectively, the second term is the CKA similarity between the multi-domain and domain-specific features.
As the original CKA similarity requires the computation of the kernel matrices over the whole datasets, which is not scalable to large datasets, we follow \cite{nguyen2020wide} and compute them over each minibatch in our training.
We refer to \cite{kornblith2019similarity,nguyen2020wide} for more details.

\subsection{Feature adaptation in meta-test}
\label{sec:adapt}
During meta-test, given a support set $\mathcal{S}=\{(\bx_i, y_i)\}_{i=1}^{|\mathcal{S}|}$ of a new learning task, we use the multi-domain model to extract features $\{f_{\phi}(\bx_i)\}_{i=1}^{|\mathcal{S}|}$ and adapt them to the target task.
To this end, we apply a linear transformation $A_{\vartheta}:\mathrm{R}^d\rightarrow \mathrm{R}^d$ with learnable parameters $\vartheta$ to the computed features, \ie $\{\bz_i\}_{i=1}^{|\mathcal{S}|} = \{A_{\vartheta} \circ f_{\phi}(\bx_i)\}_{i=1}^{|\mathcal{S}|}$ where $\vartheta\in\mathrm{R}^{d\times d}$.
Then we follow a similar pipeline to the one in \cite{dvornik2020selecting,mensink2013distance,snell2017prototypical} to build a centroid classifier by averaging the embeddings belonging to this class:
\begin{equation}
\bc_j = \frac{1}{|\mathcal{S}_j|}\sum_{\bz_i\in \mathcal{S}_j}\bz_i, \mathcal{S}_j=\{\bz_k: y_k=j\}, j=1,\dots, C
\end{equation} where $C$ is the number of classes in the support set.
Next we estimate the likelihood of a support sample $\bz$ by:
\begin{equation}\label{eq:ncc}
p(y=l|\bz) = \frac{\exp(-d(\bz, \bc_l))}{\sum_{j=1}^{C}\exp(-d(\bz, \bc_j))},
\end{equation}where $d(\bz, \bc_l)$ is the negative cosine similarity.

\begin{figure}[t!]
\begin{center}
\includegraphics[width=1.0\linewidth]{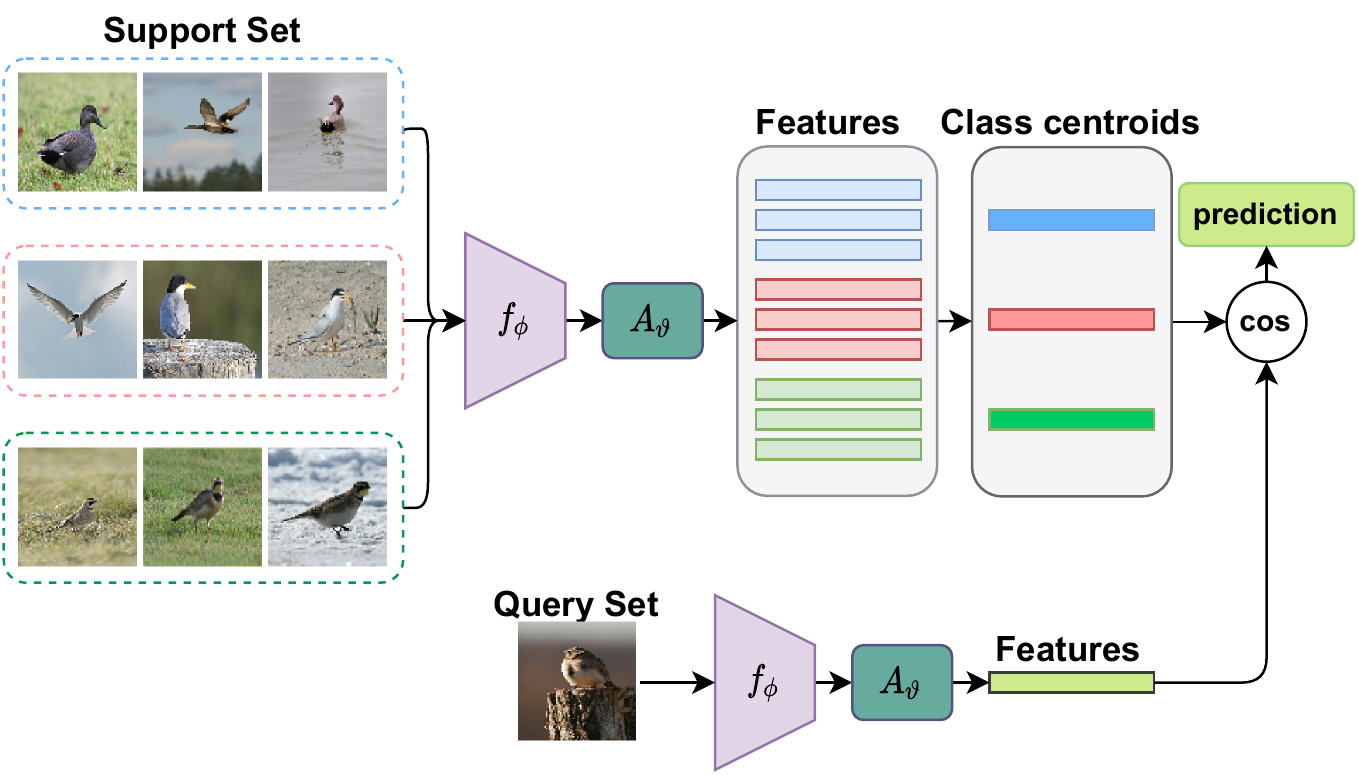}
\end{center}
\vspace{-0.45cm}
\caption{Illustration of adaptation procedure in meta-test. Given a support set and query image, our method learns to map their features to a task-specific space through a linear transformation $A_{\vartheta}$ and assign the query image to the nearest class center.}
\label{fig:meta-test}
\end{figure}

We then optimize $\vartheta$ to minimize the following objective on the support set $\mathcal{S}$:
\begin{equation}
    \label{eq:adapt}
\min_{\vartheta}\frac{1}{|\mathcal{S}|}\sum_{\bx_i, y_i \in \mathcal{S}}[log(p(y=y_i|\bx_i))].
\end{equation} Solving \cref{eq:adapt} for $\vartheta$ results in high intra-class and low inter-class similarity in the adapted space. 
We then use $\vartheta$ and \cref{eq:ncc} to predict the label of the query sample from $\mathcal{Q}$ by picking the closest centroid $\bc_j$.
Our meta-test pipeline is illustrated \cref{fig:meta-test}.

\paragraph{Discussion.} In~\cite{bateni2020improved}, Simple CNAPS uses the (squared) Mahalanobis distance between the features of class centroid and a query image, $d(\bz, \bc)=\frac{1}{2}(f_{\phi}(\bx)-\bc')^\top \mathbf{Q}^{-1} (f_{\phi}(\bx)-\bc')$ where $\mathbf{Q}$ is a covariance matrix specific to the task and class and $\bc'$ is the class centroid in the feature space (before the adaptation).
The authors show that considering the class covariance enables better adaptation of the feature extractor to the target task.
Our adaptation strategy can be seen as a generalization of the Mahalanobis distance computation.
Alternatively, assuming that $\mathbf{Q}^{-1}$ can be decomposed into a product of a lower triangular matrix and its conjugate transpose, \ie $\mathbf{Q}^{-1}=\mathbf{L} \mathbf{L}^\top$, one can first pre-transform the features by multiplication, \ie $\bz=\mathbf{L}^\top f_{\phi}(\bx)$ and then compute the distance between these features and centroids.
Similarly, we apply a linear transformation to the features but unlike \cite{bateni2020improved}, we learn its parameters $\vartheta$ by optimizing \cref{eq:adapt}.

\section{Experiments}\label{sec:exp}

\begin{table*}[ht]
	\centering
    \resizebox{1.0\textwidth}{!}
    {
		\begin{tabular}{cccccccccc}

		    \toprule
		    Test Dataset & Proto-MAML~\cite{triantafillou2019meta} & BOHB-E~\cite{saikia2020optimized} & CNAPS~\cite{requeima2019fast} & Best SDL & MDL & Simple CNAPS~\cite{bateni2020improved} & SUR~\cite{dvornik2020selecting} & URT~\cite{liu2020universal} & Ours \\
		    \midrule
		    ImageNet & $46.5\pm1.1$ & $51.9\pm1.1$ & $50.8\pm1.1$ & $55.8\pm1.0$ & $53.4\pm1.1$ & $58.4\pm1.1$ & $56.2\pm1.00$ & $56.8\pm1.1$ & ${\bf 58.8\pm1.1}$\\
		    Omniglot & $82.7\pm1.0$ & $67.6\pm1.2$ & $91.7\pm0.5$ & $93.2\pm0.5$ & $93.8\pm0.4$ &$91.6\pm0.6$ & $94.1\pm0.42$ & $94.2\pm0.4$ & ${\bf 94.5\pm0.4}$\\
		    Aircraft & $75.3\pm0.8$ & $54.1\pm0.9$ & $83.7\pm0.6$ & $85.7\pm0.5$ & $86.6\pm0.5$ & $82.0\pm0.7$ & $85.5\pm0.54$ & $85.8\pm0.5$ &  ${\bf 89.4\pm0.4}$\\
		    Birds & $69.9\pm1.0$ & $70.7\pm0.9$ & $73.6\pm0.9$ & $71.2\pm0.9$ & $78.6\pm0.8$ & $74.8\pm0.9$ & $71.0\pm1.00$ & $76.2\pm0.8$ & ${\bf 80.7\pm0.8}$\\
		    Textures & $68.3\pm0.8$ & $68.3\pm0.8$ & $59.5\pm0.7$ & $73.0\pm0.6$ & $71.4\pm0.7$ & $68.8\pm0.9$ & $71.0\pm0.80$ & $71.6\pm0.7$ &  ${\bf 77.2\pm0.7}$\\
		    Quick Draw & $66.8\pm0.9$ & $50.3\pm1.0$ & $74.7\pm0.8$ & ${\bf 82.8\pm0.6}$ &$81.5\pm0.6$ & $76.5\pm0.8$ & $81.8\pm0.57$ & $82.4\pm0.6$ & $82.5\pm0.6$\\
		    Fungi & $42.0\pm1.2$ & $41.4\pm1.1$ & $50.2\pm1.1$ & $65.8\pm0.9$ & $61.9\pm1.0$ & $46.6\pm1.0$ & $64.3\pm0.95$ & $64.0\pm1.0$ & ${\bf 68.1\pm0.9}$\\
		    VGG Flower & $88.7\pm0.7$ & $87.3\pm0.6$ & $88.9\pm0.5$ & $87.0\pm0.6$ & $88.7\pm0.6$ & $90.5\pm0.5$ & $82.9\pm0.78$ & $87.9\pm0.6$ &  ${\bf 92.0\pm0.5}$\\
		    \midrule
		    Traffic Sign & $52.4\pm1.1$ & $51.8\pm1.0$ & $56.5\pm1.1$ & $47.4\pm1.1$ & $51.0\pm1.0$ & $57.2\pm1.0$ & $51.0\pm1.11$ & $48.3\pm1.1$ & ${\bf 63.3\pm1.2}$\\
		    MSCOCO & $41.7\pm1.1$ & $48.0\pm1.0$ & $39.4\pm1.0$ & $53.5\pm1.0$ & $49.7\pm1.1$ & $48.9\pm1.1$ & $52.0\pm1.09$ & $51.5\pm1.1$ & ${\bf 57.3\pm1.0}$\\
		    MNIST & - & - & - & $89.8\pm0.5$  & $94.4\pm0.3$ & $94.6\pm0.4$ & $94.3\pm0.41$ & $90.6\pm0.5$ & ${\bf 94.7\pm0.4}$\\
		    CIFAR-10 & - & - & - & $67.3\pm0.8$ & $66.7\pm0.8$  & ${\bf 74.9\pm0.7}$ & $66.5\pm0.89$ & $67.0\pm0.8$ & $74.2\pm0.8$  \\
		    CIFAR-100 & - & - & - & $56.6\pm0.9$ & $53.6\pm1.0$ & $61.3\pm1.1$ & $56.9\pm1.10$ & $57.3\pm1.0$ & ${\bf 63.6\pm1.0}$ \\
		    \midrule
		    Average Rank & 7.8 & 8.1 & 6.6 & 4.8 & 4.6 & 5.2 & 5.0 & 4.4 & 1.3 \\
			\bottomrule
		\end{tabular}%
			}
		\vspace{-0.15cm}
		\caption{Comparison to baselines and state-of-the-art methods on Meta-Dataset. Mean accuracy, 95\% confidence interval are reported. The first eight datasets are seen during training and the last five datasets are unseen and used for test only. Average rank is computed according to first 10 datasets as some methods do not report results on last three datasets.}
		\label{tab:currmethod}
\end{table*}%

Here we first describe the benchmarks, implementation details and competing methods. 
Then we rigorously compare our method to the state-of-the-art and also study each proposed component in an ablation. We also analyze our method qualitatively. Finally we evaluate our method in a global retrieval task to further evaluate the learned feature representations in few-shot classification task.

\subsection{Experimental setup}
\paragraph{Dataset.} Meta-Dataset~\cite{triantafillou2019meta} is a few-shot classification benchmark that initially consisted of ten datasets: ILSVRC\_2012~\cite{russakovsky2015imagenet} (ImageNet),  Omniglot~\cite{Lake1332}, FGVC-Aircraft~\cite{maji2013fine} (Aircraft), CUB-200-2011~\cite{wah2011caltech} (Birds), Describable Textures~\cite{cimpoi2014describing} (DTD), QuickDraw~\cite{jongejan2016quick}, FGVCx Fungi~\cite{brigit2018fungi} (Fungi), VGG Flower~\cite{nilsback2008automated} (Flower), Traffic Signs~\cite{houben2013detection} and MSCOCO~\cite{lin2014microsoft} then further expanded with MNIST~\cite{lecun1998gradient}, CIFAR-10~\cite{krizhevsky2009learning} and CIFAR-100~\cite{krizhevsky2009learning}. 
We follow the standard procedure and use the first eight datasets for meta-training, in which each dataset is further divided into train, validation and test set with disjoint classes. 
The evaluation within these datasets is used to measure the generalization ability in the seen domains.
The rest five datasets are reserved as unseen domain for meta-test for measuring the cross-domain generalization ability.  

\paragraph{Implementation details.}
We use PyTorch~\cite{NEURIPS2019_9015} library to implement our method. 
In all experiments we build our method on ResNet-18~\cite{he2016deep} backbone for both single-domain and multi-domain networks.
In the multi-domain network, we share all the layers but the last classifier across the domains.
For training single-domain models, we strictly follow the training protocol in~\cite{dvornik2020selecting}, use a SGD optimizer with a momentum and the cosine annealing learning scheduler with the same hyperparameters.
For our multi-domain network, we use the same optimizer and scheduler as before, train it for 240,000 iterations.
We set $\lambda^f$ and $\lambda^p$ as 4 for ImageNet and 1 for other datasets and use early-stopping based on cross-validation over the validations sets of 8 training datasets.
We refer to supplementary for more details.

\paragraph{Baselines and compared methods.}
First we compare our method to our own baselines, i) the best single-domain model (Best SDL) where we use each single-domain network as the feature extractor and test it for few-shot classification in each dataset and pick the best performing model. (See supplementary for the complete results) This involves evaluating 8 single-domain networks on 13 datasets, serves a very competitive baseline, ii) the vanilla multi-domain learning baseline (MDL) that is learning by optimizing \cref{eq:mtl} without the proposed distillation method.
As additional baseline, we include the best performing method in~\cite{triantafillou2019meta}, \ie Proto-MAML~\cite{triantafillou2019meta}, and as well as the state-of-the-art methods, OHB-E~\cite{saikia2020optimized}, CNAPS~\cite{requeima2019fast}, SUR~\cite{dvornik2020selecting}, URT~\cite{liu2020universal}, and the Simple CNAPS~\cite{bateni2020improved}\footnote{Results of Proto-MAML~\cite{triantafillou2019meta}, BOHB-E~\cite{saikia2020optimized}, and CNAPS~\cite{requeima2019fast} are obtained from \href{https://github.com/google-research/meta-dataset}{Meta-Dataset}.}.
For evaluation, we follow the standard protocol in~\cite{triantafillou2019meta}, randomly sample 600 tasks for each dataset, and report average accuracy and 95\% confidence score in all experiments. 
We reproduce results by training and evaluating SUR~\cite{dvornik2020selecting}, URT~\cite{liu2020universal}, and Simple CNAPS~\cite{bateni2020improved} using their code for fair comparison as recommended by \href{https://github.com/google-research/meta-dataset}{Meta-Dataset}.

\begin{table}[ht]
	\centering
    \resizebox{1.0\textwidth}{!}
    {
		\begin{tabular}{ccccc|cccc}
			& \multicolumn{4}{c}{Varying-Way Five-Shot} & \multicolumn{4}{c}{Five-Way One-Shot}\\
		    \toprule
		    \multirow{2}{*}{Test Dataset} & Simple & SUR & URT & \multirow{2}{*}{Ours} & Simple & SUR & URT & \multirow{2}{*}{Ours}\\
		    &  CNAPS~\cite{bateni2020improved}  & \cite{dvornik2020selecting} & \cite{liu2020universal} & & CNAPS~\cite{bateni2020improved} & \cite{dvornik2020selecting}& \cite{liu2020universal} & \\
		    \midrule
		    ImageNet     & $47.2$& $46.7$& $48.6$& ${\bf 49.5}$ & $42.6$& $40.7$& $47.4$& ${\bf 48.1}$\\
		    Omniglot     & $95.1$& $95.8$& $96.0$& ${\bf 96.4}$ & $93.1$& $93.0$& $95.6$& ${\bf 96.1}$\\
		    Aircraft     & $74.6$& $82.1$& $81.2$& ${\bf 84.4}$ & $65.8$&  $67.1$& $77.9$& ${\bf 81.6}$\\
		    Birds        & $69.6$& $62.8$& $71.2$& ${\bf 75.6}$ &  $67.9$& $59.2$& $70.9$& ${\bf 75.7}$\\
		    Textures     & $57.5$& $60.2$& $65.2$& ${\bf 65.7}$ &$42.2$& $42.5$& $49.4$& ${\bf 52.4}$\\
		    Quick Draw   & $70.9$& $79.0$& ${\bf 79.2}$& $78.3$ &$70.5$& ${\bf 79.8}$& $79.6$& $79.4$\\
		    Fungi        & $50.3$& $66.5$& $66.9$& ${\bf 68.1}$ &$58.3$& $64.8$& $71.0$& ${\bf 73.7}$\\
		    VGG Flower   & $86.5$& $76.9$& $82.4$& ${\bf 86.3}$ &$79.9$& $65.0$& $72.7$& ${\bf 80.0}$\\
		    \midrule
		    Traffic Sign & $55.2$& $44.9$& $45.1$& ${\bf 57.6}$ &$55.3$& $44.6$& $52.7$& ${\bf 56.4}$\\
		    MSCOCO       & $49.2$& $48.1$& $52.3$& ${\bf 54.7}$ & $48.8$& $47.8$& $56.9$& ${\bf 58.5}$\\
		    MNIST        & $88.9$& ${\bf 90.1}$& $86.5$& $89.4$ & ${\bf 80.1}$& $77.1$& $75.6$& $78.9$\\
		    CIFAR-10     & ${\bf 66.1}$& $50.3$& $61.4$& $64.6$ & $50.3$& $35.8$& $47.3$& ${\bf 53.2}$\\
		    CIFAR-100    & $53.8$& $46.4$& $52.5$& ${\bf 54.9}$ & $53.8$& $42.9$& $54.9$& ${\bf 61.3}$\\
		    \midrule
		    Average Rank & 3.0 & 3.0 & 2.5 & 1.5 & 2.8 & 3.5 & 2.3 & 1.3 \\
			\bottomrule
		\end{tabular}%
			}
		\vspace{-0.15cm}
		\caption{Results of Varying-Way Five-Shot and Five-Way One-Shot settings. Mean accuracies are reported and the results with confidence interval are shown in the supplementary.}
		\label{tab:fixedshot}
\end{table}%

\subsection{Results}
As in Meta-Dataset~\cite{triantafillou2019meta}, we sample each task with varying number of ways and shots and report the results in \cref{tab:currmethod}.
Our method outperforms the state-of-the-art methods  in seven out of eight seen datasets and four out of five unseen datasets. 
We also compute average rank as recommended in~\cite{triantafillou2019meta}, our method ranks 1.3 in average and the state-of-the-art methods SUR and URT rank 5.0 and 4.4, respectively.
More specifically, we obtain significant better results than the second best approach  on Aircraft (+2.8), Birds (+2.1), Texture (+4.2), and VGG Flower (+1.5) for seen domains and Traffic Sign (+6.1)\footnote{The accuracy of all methods on Traffic Sign is different from the one in the original papers as one bug has been fixed in Meta-Dataset repository. See \url{https://github.com/google-research/meta-dataset/issues/54} for more details.} and MSCOCO (+3.8). 
The results show that jointly learning a single set of representations provides better generalization ability than fusing the ones from multiple single-domain feature extractors as done in SUR and URT.
Notably, our method requires less parameters and less computations to run during inference than SUR and URT, as it runs only one universal network to extract features, while both SUR and URT need to pass the query set to multiple single-domain network.

\begin{table}[th]
	\centering
    \resizebox{1.0\textwidth}{!}
    {
		\begin{tabular}{cccccc}

		    \toprule
		    Test Dataset & L2 & COSINE & CKA & KL & CKA + KL \\
		    \midrule
		    ImageNet & $55.7\pm1.1$ & $57.0\pm1.1$& ${\bf 59.0\pm1.0}$ & $57.0\pm1.1$& $58.8\pm1.1$ \\
		    Omniglot & $94.0\pm0.4$ & $94.1\pm0.4$& ${\bf 94.7\pm0.4}$ & $94.5\pm0.4$ & $94.5\pm0.4$\\
		    Aircraft & $87.4\pm0.5$ & $88.3\pm0.5$& $88.9\pm0.5$ & $89.3\pm0.4$ & ${\bf 89.4\pm0.4}$\\
		    Birds & $78.5\pm0.7$ & $77.5\pm0.8$& $80.4\pm0.7$ & $78.6\pm0.8$ & ${\bf 80.7\pm0.8}$\\
		    Textures & $72.8\pm0.6$ & $73.2\pm0.7$& $74.5\pm0.7$ & $73.3\pm0.7$ & ${\bf 77.2\pm0.7}$\\
		    Quick Draw & $81.2\pm0.6$ & $80.8\pm0.6$& $81.9\pm0.6$ & $81.6\pm0.6$ & ${\bf 82.5\pm0.6}$\\
		    Fungi & $65.7\pm0.9$ & $65.9\pm0.9$& $66.4\pm0.9$ & $67.6\pm0.9$ & ${\bf 68.1\pm0.9}$\\
		    VGG Flower & $87.5\pm0.6$ & $85.0\pm0.6$& $91.3\pm0.5$ & $89.6\pm0.5$ & ${\bf 92.0\pm0.5}$\\
		    \midrule
		    Traffic Sign & $61.6\pm1.1$ & $59.5\pm1.1$& $63.2\pm1.1$ & $62.5\pm1.2$ & ${\bf 63.3\pm1.2}$\\
		    MSCOCO & $53.4\pm1.0$ & $53.8\pm1.1$& $56.6\pm1.0$ & $55.6\pm1.1$ & ${\bf 57.3\pm1.0}$\\
		    MNIST & $94.7\pm0.3$ & $93.2\pm0.5$& $94.7\pm0.4$ & ${\bf 95.3\pm0.4}$ & $94.7\pm0.4$\\
		    CIFAR-10 & $71.1\pm0.8$ & $68.1\pm0.8$& $73.8\pm0.7$ & $72.9\pm0.8$ & ${\bf 74.2\pm0.8}$\\
		    CIFAR-100 & $59.1\pm1.0$ & $58.1\pm1.0$& $62.1\pm1.0$ & $60.8\pm1.0$ & ${\bf 63.6\pm1.0}$\\
			\bottomrule
		\end{tabular}%
			}
		\vspace{-0.25cm}
		\caption{Comparison of loss functions for knowledge distillation. Mean accuracy, 95\% confidence interval are reported. L2 denotes L2 loss between two feature representations. COSINE represents negative cosine similarity function. KL means KL divergence loss function on the network predictions. All results are obtained with feature adaptation during meta-test stage.}
		\label{tab:lossf}
\end{table}%

We also see that our method outperforms two strong baselines, Best SDL and MDL in all datasets except in QuickDraw.
This indicates that i) universal representations are superior to the single-domain ones while generalizing to new tasks in both seen and unseen domains, while requiring significantly less number of parameters (1 vs 8 neural networks), ii) our distillation strategy is essential to obtain good multi-domain representations.
While MDL outperforms the best SDL in certain domains by transferring representations across them, its performance is lower in other domains than SDL, possibly due to negative transfer across the significantly diverse domains.
Surprisingly, MDL achieves the third best in average rank, indicating the benefit of multi-domain representations.

\begin{table}[t]
	\centering
	
    \resizebox{1.0\textwidth}{!}
    {
		\begin{tabular}{cccccc}

		    \toprule
		    Test Dataset & NCC & NCC+MD & LR & SVM  & Ours\\
		    \midrule
		    ImageNet & $57.0\pm1.1$& $53.9\pm1.0$ & $56.0\pm1.1$ & $54.5\pm1.1$ & ${\bf 58.8\pm1.1}$ \\
		    Omniglot & $94.4\pm0.4$& $93.8\pm0.5$ & $93.7\pm0.5$ & $94.3\pm0.5$ & ${\bf 94.5\pm0.4}$ \\
		    Aircraft & $88.0\pm0.5$& $87.6\pm0.5$ & $88.3\pm0.6$ & $87.7\pm0.5$ & ${\bf 89.4\pm0.4}$ \\
		    Birds & $80.3\pm0.7$& $78.3\pm0.7$ & $79.7\pm0.8$ & $78.1\pm0.8$ & ${\bf 80.7\pm0.8}$ \\
		    Textures & $74.6\pm0.7$& $73.7\pm0.7$ & $74.7\pm0.7$ & $73.8\pm0.8$ & ${\bf 77.2\pm0.7}$ \\
		    Quick Draw & $81.8\pm0.6$& $80.9\pm0.7$ & $80.0\pm0.7$ & $80.0\pm0.6$ & ${\bf 82.5\pm0.6}$ \\
		    Fungi & $66.2\pm0.9$& $57.7\pm0.9$ & $62.1\pm0.8$ & $58.5\pm0.9$ & ${\bf 68.1\pm0.9}$ \\
		    VGG Flower & $91.5\pm0.5$& $89.7\pm0.6$ & $91.1\pm0.5$ & $91.4\pm0.6$ & ${\bf 92.0\pm0.5}$ \\
		    \midrule
		    Traffic Sign & $49.8\pm1.1$& $62.2\pm1.1$ & $59.7\pm1.1$ & ${\bf 65.7\pm1.2}$ & $63.3\pm1.2$ \\
		    MSCOCO & $54.1\pm1.0$& $48.5\pm1.0$ & $51.2\pm1.1$ & $50.5\pm1.0$ & ${\bf 57.3\pm1.0}$ \\
		    MNIST & $91.1\pm0.4$& $95.1\pm0.4$ & $93.5\pm0.5$ & ${\bf 95.4\pm0.4}$ & $94.7\pm0.4$ \\
		    CIFAR-10 & $70.6\pm0.7$& $68.9\pm0.8$ & $73.1\pm0.8$ & $72.0\pm0.8$ & ${\bf 74.2\pm0.8}$ \\
		    CIFAR-100 & $59.1\pm1.0$& $60.0\pm0.9$ & $60.1\pm1.1$ & $60.5\pm1.1$ & ${\bf 63.6\pm1.0}$ \\
			\bottomrule
		\end{tabular}%
			}
		\vspace{-0.25cm}
		\caption{Comparison of different classifiers that are incorporated to our method during meta-test stage. NCC, MD, LR, SVM denote nearest center classifier, Mahalanobis distance, logistic regression, support vector machines respectively.}
		\label{tab:testad}
\end{table}%

\begin{table*}[ht]
	\centering
    \resizebox{1.0\textwidth}{!}
    {
		\begin{tabular}{c|cc|cc|cc|cc|cc|cc|cc|cc||cc|cc|cc|cc|cc}

		    \toprule
		    Test Dataset & \multicolumn{2}{c|}{ImageNet} & \multicolumn{2}{c|}{Omniglot} & \multicolumn{2}{c|}{Aircraft} & \multicolumn{2}{c|}{Birds} & \multicolumn{2}{c|}{Textures} & \multicolumn{2}{c|}{Quick Draw} & \multicolumn{2}{c|}{Fungi} & \multicolumn{2}{c||}{VGG Flower} & \multicolumn{2}{c|}{Traffic Sign} & \multicolumn{2}{c|}{MSCOCO} & \multicolumn{2}{c|}{MNIST} & \multicolumn{2}{c|}{CIFAR-10} & \multicolumn{2}{c}{CIFAR-100}\\
		    \midrule
		    Recall@$k$ & 1 & 2 & 1 & 2 & 1 & 2 & 1 & 2 & 1 & 2 & 1 & 2 & 1 & 2 & 1 & 2 & 1 & 2 & 1 & 2 & 1 & 2 & 1 & 2 & 1 & 2 \\
		    \midrule
		    Sum & $22.1$ & $30.3$ & $84.7$ & $91.8$ & $69.7$ & $80.7$ & $45.9$ & $59.7$ & $66.3$ & $78.2$ & $77.4$ & $84.3$ & $31.9$ & $42.9$ & $85.1$ & $92.1$ & $94.6$ & $97.2$ & $62.6$ & $71.2$ & $98.3$ & $99.2$ & $54.0$ & $68.9$ & $27.8$ & $37.4$ \\
		    Concate & $20.2$ & $28.0$ & $84.4$ & $91.5$ & $44.3$ & $58.1$ & $35.5$ & $48.8$ & $68.8$ & $78.2$ & $73.0$ & $80.8$ & $30.7$ & $40.4$ & $83.4$ & $91.3$ & ${\bf 95.1}$ & ${\bf 97.3}$ & $60.7$ & $69.8$ & ${\bf 98.7}$ & ${\bf 99.3}$ & $49.7$ & $65.3$ & $25.4$ & $34.6$ \\
		    MDL & $29.8$ & $39.6$ & ${\bf 89.8}$ & ${\bf 94.3}$ & $80.3$ & $87.1$ & $63.2$ & $75.9$ & $67.0$ & $77.1$ & $79.5$ & $85.4$ & $40.2$ & $51.7$ & $86.9$ & $93.3$ & $89.5$ & $94.1$ & $63.6$ & $72.6$ & $97.6$ & $98.8$ & $58.9$ & $72.9$ & $31.6$ & $42.0$ \\
		    Simple CNAPS~\cite{bateni2020improved} & $34.0$ & $43.8$ & $84.9$ & $91.6$ & $70.5$ & $82.5$ & $55.9$ & $70.5$ & $64.8$ & $76.9$ & $75.3$ & $83.0$ & $29.1$ & $39.0$ & $88.1$ & $94.1$ & $79.9$ & $86.9$ & $65.2$ & $73.8$ & $97.5$ & $98.8$ & ${\bf 66.2}$ & ${\bf 79.3}$ & $33.2$ & $44.2$ \\
		    Ours & ${\bf 36.1}$ & ${\bf 46.2}$ & $89.7$ & ${\bf 94.3}$ & ${\bf 83.3}$ & ${\bf 90.4}$ & ${\bf 66.7}$ & ${\bf 78.9}$ & ${\bf 70.2}$ & ${\bf 80.8}$ & ${\bf 79.9}$ & ${\bf 86.5}$ & ${\bf 44.5}$ & ${\bf 56.2}$ & ${\bf 90.0}$ & ${\bf 94.6}$ & $87.9$ & $93.0$ & ${\bf 67.4}$ & ${\bf 76.3}$ & $97.0$ & $98.4$ & $62.1$ & $76.5$ & ${\bf 35.1}$ & ${\bf 46.1}$ \\
			\bottomrule
		\end{tabular}%
			}
		\vspace{-0.25cm}
		\caption{Global retrieval performance on Meta-Dataset. In addition to few-shot learning experiments, we evaluate our method in a non-episodic retrieval task to further compare the generalization ability of our universal representations. }
		\label{tab:recall}
\end{table*}%

\begin{figure*}
\begin{center}
\includegraphics[width=1.0\linewidth]{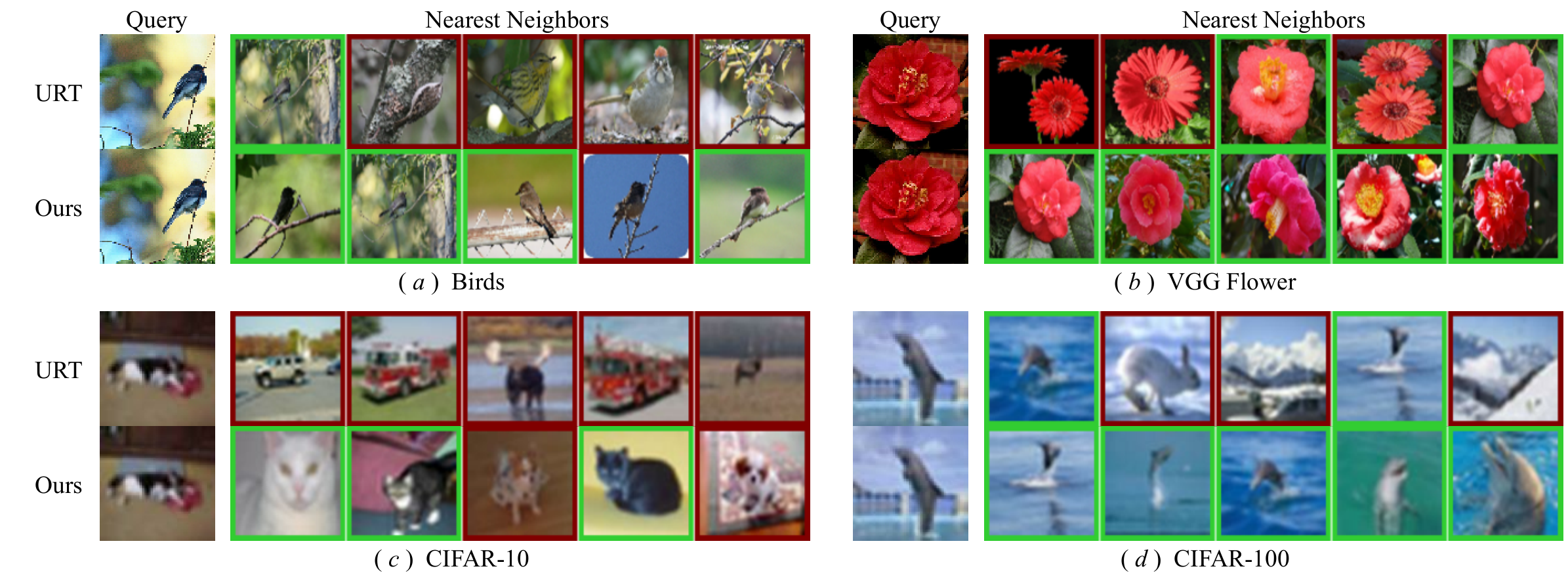}
\end{center}
\vspace{-0.45cm}
\caption{Qualitative comparison to URT in four datasets. Green and red colors indicate correct and false predictions respectively.}
\label{fig:visual}
\end{figure*}

\subsection{Further results}
\paragraph{Varying-way five-shot setting.}
After reporting results in a broad range of varying shots (\eg up to 100 shots in some extreme cases), we further analyze our method for 5-shot setting with varying number of categories.
To this end, we follow the setting in \cite{doersch2020crosstransformers}, compare our method to the best three state-of-the-art methods including Simple CNAPS, SUR and URT. In this setting, we sample a varying number of ways in Meta-Dataset the same as the standard setting but a fixed number of shots to form balanced support and query sets. 
As shown in Table~\ref{tab:fixedshot}, overall performance for all methods decreases in most datasets compared to results in Table~\ref{tab:currmethod} indicating that this is a more challenging setting. It is due to that five-shot setting samples much less support images than the standard setting.
The ranking of different methods change slightly. The top-2 methods remain the same, while both Simple CNAPS and SUR obtain 3.0 average rank. SUR performs the best on MNIST, Simple CNAPS outperforms others on CIFAR-100 and URT is top-1 on Quick Draw. 
Ours still achieves significant better performance than other methods on the rest ten datasets.

\paragraph{Results in five-way one-shot setting.}
Next we test an extremely challenging five-way one-shot setting on Meta-Dataset.
For each task, only one image per class is seen as support set.
This setting is often used in evaluating different methods in a single domain~\cite{Lake1332,ren2018meta,vinyals2016matching}, while we adopt it for multiple domains.
As shown in Table~\ref{tab:fixedshot}, our method achieves consistent gain as observed in previous two settings, which validates the importance of good universal representations in case of limited labeled samples in meta-test. Interestingly, Simple CNAPS achieves better rank than SUR in this setting, which is opposite in previous settings.

\subsection{Analyses}

Here we conduct an ablation study on different components in our framework by varying the loss function for the distillation, classifier type in meta-test.

\paragraph{Different distillation loss functions.}
First we study different distillation loss functions, including L2 loss, cosine distance, KL divergence and CKA for learning the multi-domain networks and report their performances in \cref{tab:lossf}.
While we apply KL divergence loss to match the logits of single and multi-domain networks as in \cite{hinton2015distilling}, the other loss functions are used to match the internal representations (features that are fed into classifiers) between those models.
Among the individual loss functions, the best results are obtained with either our model with CKA or KL divergence loss, while CKA outperforms KL divergence in the most domains.
Although the features are first aligned with an adapter, L2 and cosine loss functions are not sufficient to match features from very diverse domains and further aligning features with CKA is crucial.
Note that here L2 baselines corresponds to the method of \cite{li2020knowledge}.
Finally, combining CKA with KL divergence gives the best performance over the multi-domain models that are trained with the individual loss functions.

\paragraph{Different classifiers in meta-test.}
Next we evaluate the proposed adaptive mapping strategy with the nearest neighbor classifier (NCC), described in \cref{sec:adapt}, to different parametric including Support Vector Machines (SVM), Logistic Regression (LR) as in \cite{tian2020rethinking} and non-parametric classifiers including NCC without the adaptive mapping and NCC with Mahalanobis Distance (NCC+MD) in \cite{bateni2020improved} in \cref{tab:testad}.
For non-parametric classifiers, NCC performs best in unseen domains when used with Mahalanobis distance.
The parametric classifiers, SVM and LR that are trained on the limited support set obtain very competitive results and outperform the non-parametric ones in most domains.
Our method, which combines the benefit of parametric and non-parametric classifiers, outperforms SVM, LR and NCC+MD in most seen datasets, while achieves worse in some unseen domains like Traffic Sign and MNIST.

\paragraph{Qualitative results.}
We qualitatively analyze our method and compare it to URT~\cite{liu2020universal} in \cref{fig:visual} by illustrating the nearest neighbors in four different datasets given a query image (see supplementary for more examples).
It is clear that our method produces more correct neighbors than URT. URT retrieves images with more similar colors, shapes and backgrounds, while our method is able to retrieve semantically similar images. It again suggests that our method is able to learn more useful and general representations.

\subsection{Global retrieval}
Here we go beyond the few-shot classification experiments and evaluate the generalization ability of our representations that are learned in the multi-domain network in a retrieval task, inspired from metric learning literature~\cite{oh2016deep,yu2019learning}.
To this end, for each test image, we find the nearest  images in entire test set in the feature space and test whether they correspond to the same category.
For evaluation metric, we use Recall@$k$ which considers the predictions with one of the $k$ closest neighbors with the same label as positive.
In \cref{tab:recall}, we compare our method with Simple CNAPS in Recall@1 and Recall@2 (see supplementary for more results). URT and SUR require adaption using support set and no such adaptation in retrieval task is possible, we replace them with two baselines that concatenate or sum features from multiple domain-specific networks. 
Our method achieves the best performance in ten out of thirteen domains with significant gains in Aircraft, Birds, Textures and Fungi.
This strongly suggests that our multi-domain representations are the key to the success of our method in the previous few-shot classification tasks.

\section{Conclusion}\label{sec:con}
In this work, we demonstrate that learning a single set of universal representations integrated with a feature refining step achieves state-of-the-art performance in the recent Meta-Dataset benchmark.
To this end, we propose to optimize the parameters of a deep neural network simultaneously over multiple domains by aligning its features with multiple single-domain networks through linear adapters and a loss function that is inspired from CKA.
We show that the universal features can be further refined from few examples to unseen tasks by learning a transformation in a similar spirit to distance learning.
Our method outperforms the state-of-the-art techniques while using less number of parameters and being more computationally efficient than other multi-domain techniques.

{\small
\bibliographystyle{ieee_fullname}
\bibliography{ref}
}

\newpage
\appendix

\section{Implementation details}

In all experiments we build our method on ResNet-18~\cite{he2016deep} backbone for both single-domain and multi-domain networks.

\subsection{Training details of single-domain models}\label{appsec:sdl}
We train one ResNet-18 model for each training dataset. For optimization, we follow the training protocol in \cite{dvornik2020selecting}. Specifically, we use SGD optimizer and cosine annealing for all experiments with a momentum of 0.9 and a weight decay of $7\times 10^{-4}$. The learning rate, batch size, annealing frequency, maximum number of iterations are shown in \cref{apptab:hyperparams}. To regularize training, we also use the exact same data augmentations as in \cite{dvornik2020selecting}, \eg random crops and random color augmentations.

\begin{table}[ht]
	\centering
    \resizebox{1\textwidth}{!}
    {
		\begin{tabular}{ccccc}
		    \toprule
		    Dataset & learning rate & batch size & annealing freq. & max. iter. \\
		    \midrule
		    ImageNet & $3\times 10^{-2}$ & 64 & 48,000 & 480,000\\
		    Omniglot & $3\times 10^{-2}$ & 16 & 3000 & 50,000\\
		    Aircraft & $3\times 10^{-2}$ & 8 & 3000 & 50,000\\
		    Birds & $3\times 10^{-2}$ & 16 & 3000 & 50,000\\
		    Textures & $3\times 10^{-2}$ & 32 & 1500 & 50,000\\
		    Quick Draw & $1\times 10^{-2}$ & 64 & 48,000 & 480,000\\
		    Fungi & $3\times 10^{-2}$ & 32 & 15,000 & 480,000\\
		    VGG Flower & $3\times 10^{-2}$ & 8 & 1500 & 50,000\\
			\bottomrule
		\end{tabular}%
			}
		\vspace{-0.25cm}
		\caption{Training hyper-parameters of single domain learning.}
		\label{apptab:hyperparams}
\end{table}%

\subsection{Training details of our method}

In the multi-domain network, we share all the layers but the last classifier across the domains.
To train the multi-domain network, we use the same optimizer with a weight decay of $7\times 10^{-4}$ and a scheduler as single domain learning model for learning 240,000 iterations. The learning rate is 0.03 and the annealing frequency is 48,000. Similar to~\cite{triantafillou2019meta} that the training episodes have 50\% probability coming from the ImageNet data source, each training batch for our multi-domain network consists of 50\% data coming from ImageNet. In other words. The batch size for ImageNet is $64\times 7$ and is $64$ for the other 7 datasets.

We set $\lambda^f$ and $\lambda^p$ as 4 for ImageNet and 1 for other datasets, respectively. And we linearly anneal $\lambda$ by $\lambda \leftarrow \lambda \times (1 - \frac{t}{T})$, where, $t$ is the current iteration and $T$ is the total number of iterations to anneal $\lambda$ to zero. Here, $T=k\times (anneal.~freq.)$, where $anneal.~freq$. is 48, 000 in this work. We search the $k=\{1, 2, 3, 4, 5\}$ based on cross-validation over the validation sets of 8 training datasets and $k$ is 5 (\ie $T=240,000$) for ImageNet, is 2 for Omniglot, Quick Draw, Fungi and is 1 for other datasets. For all experiments, early-stopping is performed based on cross-validation over the validations sets of 8 training datasets.

For the optimization of feature adaptation during meta-test stage, we initialize $\vartheta$ as an indentity matrix, which allows the NCC to use the original features produced by our universal network and optimize the adaptor $\vartheta$ from a good start point. Similar to the optimization in~\cite{dvornik2020selecting}, we optimize $\vartheta$ for 40 iterations using Adadelta~\cite{zeiler2012adadelta} as optimizer with a learning rate of 0.1 for first eight datasets and 1 for the last five datasets.

\section{More results}
In this section, we first evaluate each single-domain model for few-shot classification on each test dataset. Then we evaluate the effect of the adaptors for aligning features in knowledge distillation. Then we show complete results on varying-way five-shot and five-way one-shot settings. Finally more qualitative results and global retrieval results are reported. 
\subsection{Complete results of single domain learning}

\begin{table*}[t]
	\centering
	\vspace{-0.2cm}
    \resizebox{0.8\textwidth}{!}
    {
		\begin{tabular}{c|cccccccc}

		    \toprule
		    \diagbox{Test Dataset}{Train Dataset} & ImageNet & Omniglot & Aircraft & Birds & Textures & Quick Draw & Fungi & Vgg Flower\\
		    \midrule
		    ImageNet & ${\bf 55.8\pm1.0}$& $17.1\pm0.6$& $21.7\pm0.7$& $25.4\pm0.8$& $24.2\pm0.8$& $24.1\pm0.8$& $32.9\pm0.9$& $25.0\pm0.8$ \\
		    Omniglot & $67.4\pm1.2$& ${\bf 93.2\pm0.5}$& $58.2\pm1.2$& $58.7\pm1.4$& $57.3\pm1.4$& $78.4\pm1.0$& $57.6\pm1.3$& $54.6\pm1.3$ \\
		    Aircraft & $49.5\pm0.9$& $16.8\pm0.5$& ${\bf 85.7\pm0.5}$& $31.4\pm0.8$& $26.0\pm0.7$& $23.8\pm0.6$& $31.0\pm0.7$& $24.6\pm0.6$ \\
		    Birds & ${\bf 71.2\pm0.9}$& $13.0\pm0.6$& $19.9\pm0.7$& $65.0\pm0.9$& $19.6\pm0.7$& $16.7\pm0.7$& $42.8\pm1.0$& $28.9\pm0.8$ \\
		    Textures & ${\bf 73.0\pm0.6}$& $25.0\pm0.5$& $38.6\pm0.7$& $42.2\pm0.7$& $54.9\pm0.7$& $38.6\pm0.6$& $54.1\pm0.7$& $42.3\pm0.7$ \\
		    Quick Draw & $53.9\pm1.0$& $51.0\pm1.0$& $38.8\pm1.0$& $38.2\pm1.0$& $36.8\pm0.9$& ${\bf 82.8\pm0.6}$& $37.7\pm0.9$& $39.7\pm1.0$ \\
		    Fungi & $41.6\pm1.0$&  $9.1\pm0.5$& $14.9\pm0.7$& $25.5\pm0.8$& $15.6\pm0.7$& $12.5\pm0.6$& ${\bf 65.8\pm0.9}$& $23.3\pm0.8$ \\
		    VGG Flower & ${\bf 87.0\pm0.6}$& $23.8\pm0.6$& $45.5\pm0.8$& $62.9\pm0.8$& $44.4\pm0.8$& $33.4\pm0.7$& $79.6\pm0.7$& $78.3\pm0.7$ \\
		    \midrule
		    Traffic Sign & ${\bf 47.4\pm1.1}$& $15.1\pm0.7$& $30.8\pm0.9$& $31.0\pm0.9$& $38.8\pm1.1$& $31.1\pm0.9$& $28.0\pm0.9$& $30.4\pm0.9$ \\
		    MSCOCO & ${\bf 53.5\pm1.0}$& $12.9\pm0.6$& $22.5\pm0.8$& $25.1\pm0.9$& $23.7\pm0.8$& $21.3\pm0.8$& $32.5\pm1.0$& $25.7\pm0.8$ \\
		    MNIST & $78.1\pm0.7$& ${\bf 89.8\pm0.5}$& $68.0\pm0.8$& $73.0\pm0.7$& $64.5\pm0.8$& $88.2\pm0.5$& $62.2\pm0.8$& $72.1\pm0.7$ \\
		    CIFAR-10 & ${\bf 67.3\pm0.8}$& $28.5\pm0.6$& $41.2\pm0.7$& $41.8\pm0.8$& $36.9\pm0.7$& $40.0\pm0.7$& $38.8\pm0.7$& $41.3\pm0.8$ \\
		    CIFAR-100 & ${\bf 56.6\pm0.9}$& $12.3\pm0.6$& $24.3\pm0.9$& $28.8\pm0.9$& $24.2\pm0.9$& $23.4\pm0.8$& $25.2\pm0.9$& $29.1\pm1.0$ \\
			\bottomrule
		\end{tabular}%
			}
		\vspace{-0.25cm}
		\caption{Results of all single domain learning models. Mean accuracy and 95\% confidence interval are reported. The first eight datasets are seen during training and the last five datasets are unseen for test only.}
		\label{tabapp:stl}
\end{table*}%

\begin{table*}[ht]
\vspace{-0.25cm}
	\centering
	    \resizebox{0.85\textwidth}{!}
    {
		\begin{tabular}{ccccc|cccc}
			& \multicolumn{4}{c}{Five-Shot} & \multicolumn{4}{c}{Five-Way One-Shot} \\
		    \toprule
		    \multirow{2}{*}{Test Dataset} & Simple & SUR & URT & \multirow{2}{*}{Ours} & Simple & SUR & URT & \multirow{2}{*}{Ours}\\
		    &  CNAPS~\cite{bateni2020improved}  & \cite{dvornik2020selecting} & \cite{liu2020universal} & & CNAPS~\cite{bateni2020improved} & \cite{dvornik2020selecting}& \cite{liu2020universal} & \\
		    \midrule
		    ImageNet & $47.2\pm1.0$& $46.7\pm1.0$& $48.6\pm1.0$& ${\bf 49.5\pm1.0}$ & $42.6\pm0.9$& $40.7\pm1.0$& $47.4\pm1.0$& ${\bf 48.1\pm1.0}$\\
		    Omniglot & $95.1\pm0.3$& $95.8\pm0.3$& $96.0\pm0.3$& ${\bf 96.4\pm0.3}$ & $93.1\pm0.5$& $93.0\pm0.7$& $95.6\pm0.5$& ${\bf 96.1\pm0.5}$\\
		    Aircraft & $74.6\pm0.6$& $82.1\pm0.6$& $81.2\pm0.6$& ${\bf 84.4\pm0.5}$ & $65.8\pm0.9$&  $67.1\pm1.4$& $77.9\pm0.9$& ${\bf 81.6\pm0.9}$\\
		    Birds & $69.6\pm0.7$& $62.8\pm0.9$& $71.2\pm0.7$& ${\bf 75.6\pm0.6}$ &  $67.9\pm0.9$& $59.2\pm1.0$& $70.9\pm0.9$& ${\bf 75.7\pm0.9}$\\
		    Textures & $57.5\pm0.7$& $60.2\pm0.7$& $65.2\pm0.7$& ${\bf 65.7\pm0.7}$ &$42.2\pm0.8$& $42.5\pm0.8$& $49.4\pm0.9$& ${\bf 52.4\pm0.9}$\\
		    Quick Draw & $70.9\pm0.6$& $79.0\pm0.5$& ${\bf 79.2\pm0.5}$& $78.3\pm0.5$ &$70.5\pm0.9$& ${\bf 79.8\pm0.9}$& $79.6\pm0.9$& $79.4\pm0.9$\\
		    Fungi & $50.3\pm1.0$& $66.5\pm0.8$& $66.9\pm0.9$& ${\bf 68.1\pm0.8}$ & $58.3\pm1.1$& $64.8\pm1.1$& $71.0\pm1.0$& ${\bf 73.7\pm1.0}$\\
		    VGG Flower & $86.5\pm0.4$& $76.9\pm0.6$& $82.4\pm0.5$& ${\bf 86.3\pm0.5}$ & $79.9\pm0.7$& $65.0\pm1.0$& $72.7\pm0.0$& ${\bf 80.0\pm0.8}$\\
		    \midrule
		    Traffic Sign & $55.2\pm0.8$& $44.9\pm0.9$& $45.1\pm0.9$& ${\bf 57.6\pm0.8}$ & $55.3\pm0.9$& $44.6\pm0.9$& $52.7\pm0.9$& ${\bf 56.4\pm0.9}$\\
		    MSCOCO & $49.2\pm0.8$& $48.1\pm0.9$& $52.3\pm0.9$& ${\bf 54.7\pm0.8}$ & $48.8\pm0.9$& $47.8\pm1.1$& $56.9\pm1.1$& ${\bf 58.5\pm1.0}$\\
		    MNIST & $88.9\pm0.4$& ${\bf 90.1\pm0.4}$& $86.5\pm0.5$& $89.4\pm0.4$ & ${\bf 80.1\pm0.9}$& $77.1\pm0.9$& $75.6\pm0.9$& $78.9\pm0.8$\\
		    CIFAR-10 & ${\bf 66.1\pm0.7}$& $50.3\pm1.0$& $61.4\pm0.7$& $64.6\pm0.7$ & $50.3\pm0.9$& $35.8\pm0.8$& $47.3\pm0.9$& ${\bf 53.2\pm0.8}$\\
		    CIFAR-100 & $53.8\pm0.9$& $46.4\pm0.9$& $52.5\pm0.9$& ${\bf 54.9\pm0.8}$ & $53.8\pm0.9$& $42.9\pm1.0$& $54.9\pm1.1$& ${\bf 61.3\pm0.9}$\\
		    \midrule
		    Average Rank & 3.0 & 3.0 & 2.5 & 1.5 & 2.8 & 3.5 & 2.3 & 1.3 \\
			\bottomrule
		\end{tabular}%
			}
		\vspace{-0.25cm}
		\caption{Results of Five-Way One-Shot and Varying-Way Five-Shot settings. Mean accuracies are reported and the results with confidence interval are reported.}
		\label{tabapp:fixedshot}
\end{table*}%

To study the universal representation learning from multiple datasets, we train one network on each training dataset and use each single-domain network as the feature extractor and test it for few-shot classification in each dataset. This involves evaluating 8 single-domain networks on 13 datasets using Nearest Centroid Classifier (NCC). \Cref{tabapp:stl} shows the results of single domain learning models, where each column present the mean accuracy and 95\% confidence interval of a single-domain network trained on one dataset (\eg ImageNet) and evaluated on 13 test datasets. The average accuracy and 95\% confidence intervals computed over 600 few-shot tasks. The numbers in bold indicate that a method has the best accuracy per dataset.

As shown in \cref{tabapp:stl}, the feature of the ImageNet model generalizes well and achieves the best results on four out of eight seen datasets, \eg ImageNet, Birds, Texture, VGG Flower and four out of five previously unseen datasets, \eg Traffic Sign, MSCOCO, CIFAR-10, CIFAR-100. The models trained on Omniglot, Aircraft, Quick Draw, and Fungi perform the best on the corresponding datasets while the Omniglot model also generalizes well to MNIST which has the similar style images to Omniglot. We then pick the best performing model, forming the best single-domain model (Best SDL) which serves a very competitive baseline for universal representation learning.

\begin{table}[ht]
\vspace{-0.3cm}
	\centering
    \resizebox{0.8\textwidth}{!}
    {
		\begin{tabular}{cccc}

		    \toprule
		    Test Dataset & Ours (CKA w/o $A_{\theta}$) & Ours (CKA) \\
		    \midrule
		    ImageNet & $58.3\pm1.0$ & ${\bf 59.0\pm1.0}$ \\
		    Omniglot & $94.4\pm0.4$ & ${\bf 94.7\pm0.4}$ \\
		    Aircraft & ${\bf 88.9\pm0.5}$ & ${\bf 88.9\pm0.4}$ \\
		    Birds & $78.7\pm0.8$ & ${\bf 80.4\pm0.7}$ & \\
		    Textures & ${\bf 74.8\pm0.7}$ & $74.5\pm0.7$ \\
		    Quick Draw  & ${\bf 82.1\pm0.6}$ & $81.9\pm0.6$ \\
		    Fungi & $65.4\pm0.9$ & ${\bf 66.4\pm0.9}$ \\
		    VGG Flower & $87.5\pm0.6$ & ${\bf 91.3\pm0.5}$ \\
		    \midrule
		    Traffic Sign & ${\bf 63.3\pm1.1}$ & $63.2\pm1.1$ \\
		    MSCOCO & $55.3\pm1.0$ & ${\bf 56.6\pm1.0}$ \\
		    MNIST & ${\bf 94.9\pm0.4}$ & $94.7\pm0.4$ \\
		    CIFAR-10 & $73.4\pm0.7$ & ${\bf 73.8\pm0.7}$ \\
		    CIFAR-100 & $61.8\pm1.0$ & ${\bf 62.1\pm1.0}$ \\
			\bottomrule
		\end{tabular}%
			}
		\vspace{-0.25cm}
		\caption{Results of our method using CKA, CKA without adaptors (\ie $A_{\theta}$). Mean accuracy and 95\% confidence interval are reported. Here, Ours (CKA w/o $A_{\theta}$) indicates that adaptors are not applied for aligning features. All results are obtained with feature adaptation during meta-test stage.}
		\label{apptab:lossf}
\end{table}%

\begin{table*}[ht]
	\centering
    \resizebox{1\textwidth}{!}
    {
		\begin{tabular}{c|cccc|cccc|cccc|cccc|cccc|cccc|cccc|cccc}

		    \toprule
		    Test Dataset & \multicolumn{4}{c|}{ImageNet} & \multicolumn{4}{c|}{Omniglot} & \multicolumn{4}{c|}{Aircraft} & \multicolumn{4}{c|}{Birds} & \multicolumn{4}{c|}{Textures} & \multicolumn{4}{c|}{Quick Draw} & \multicolumn{4}{c|}{Fungi} & \multicolumn{4}{c}{VGG Flower} \\
		    \midrule
		    Recall@$k$ & 1 & 2 & 4 & 8 & 1 & 2 & 4 & 8 & 1 & 2 & 4 & 8 & 1 & 2 & 4 & 8 & 1 & 2 & 4 & 8 & 1 & 2 & 4 & 8 & 1 & 2 & 4 & 8 & 1 & 2 & 4 & 8 \\
		    \midrule
		    Sum & $22.1$ & $30.3$ & $39.6$ & $50.0$ & $84.7$ & $91.8$ & $95.8$ & $97.8$ & $69.7$ & $80.7$ & $88.6$ & $94.5$ & $45.9$ & $59.7$ & $72.0$ & $84.1$ & $66.3$ & $78.2$ & $87.3$ & $94.0$ & $77.4$ & $84.3$ & $89.1$ & $92.1$ & $31.9$ & $42.9$ & $54.0$ & $65.4$ & $85.1$ & $92.1$ & $96.7$ & $98.6$ \\
		    Concate & $20.2$ & $28.0$ & $36.9$ & $47.8$ & $84.4$ & $91.5$ & $95.8$ & $97.8$ & $44.3$ & $58.1$ & $71.1$ & $82.9$ & $35.5$ & $48.8$ & $62.8$ & $76.0$ & $68.8$ & $78.2$ & $87.3$ & $93.9$ & $73.0$ & $80.8$ & $86.2$ & $90.6$ & $30.7$ & $40.4$ & $51.8$ & $63.0$ & $83.4$ & $91.3$ & $95.2$ & $98.2$ \\
		    MDL & $29.8$ & $39.6$ & $49.9$ & $60.9$ & ${\bf 89.8}$ & ${\bf 94.3}$ & $96.8$ & $98.2$ & $80.3$ & $87.1$ & $92.5$ & $95.9$ & $63.2$ & $75.9$ & $84.7$ & $91.6$ & $67.0$ & $77.1$ & $85.4$ & $92.9$ & $79.5$ & $85.4$ & $89.7$ & $92.8$ & $40.2$ & $51.7$ & $63.0$ & $72.4$ & $86.9$ & $93.3$ & $96.6$ & $98.4$ \\
		    Simple CNAPS~\cite{bateni2020improved} & $34.0$ & $43.8$ & $54.4$ & $65.1$ & $84.9$ & $91.6$ & $95.5$ & $97.5$ & $70.5$ & $82.5$ & $91.3$ & $96.1$ & $55.9$ & $70.5$ & $82.0$ & $90.2$ & $64.8$ & $76.9$ & ${\bf 87.6}$ & ${\bf 94.4}$ & $75.3$ & $83.0$ & $88.0$ & $91.7$ & $29.1$ & $39.0$ & $49.6$ & $61.5$ & $88.1$ & $94.1$ & ${\bf 97.6}$ & ${\bf 99.2}$ \\
		    Ours & ${\bf 36.1}$ & ${\bf 46.2}$ & ${\bf 56.3}$ & ${\bf 66.6}$ & $89.7$ & ${\bf 94.3}$ & ${\bf 97.2}$ & ${\bf 98.3}$ & ${\bf 83.3}$ & ${\bf 90.4}$ & ${\bf 93.7}$ & ${\bf 96.3}$ & ${\bf 66.7}$ & ${\bf 78.9}$ & ${\bf 87.9}$ & ${\bf 94.1}$ & ${\bf 70.2}$ & ${\bf 80.8}$ & $87.5$ & $93.8$ & ${\bf 79.9}$ & ${\bf 86.5}$ & ${\bf 90.5}$ & ${\bf 93.2}$ & ${\bf 44.5}$ & ${\bf 56.2}$ & ${\bf 67.3}$ & ${\bf 76.4}$ & ${\bf 90.0}$ & ${\bf 94.6}$ & $97.5$ & $98.9$ \\
			\bottomrule
		\end{tabular}%
			}
		\vspace{-0.25cm}
		\caption{Global retrieval performance on Meta-Dataset (seen datasets). In addition to few-shot learning experiments, we evaluate our method in a non-episodic retrieval task to further compare the generalization ability of our universal representations. }
		\label{tab:recallseen}
\end{table*}

\begin{table*}[ht]
	\centering
    \resizebox{0.8\textwidth}{!}
    {
		\begin{tabular}{c|cccc|cccc|cccc|cccc|cccc}

		    \toprule
		    Test Dataset & \multicolumn{4}{c|}{Traffic Sign} & \multicolumn{4}{c|}{MSCOCO} & \multicolumn{4}{c|}{MNIST} & \multicolumn{4}{c|}{CIFAR-10} & \multicolumn{4}{c}{CIFAR-100} \\
		    \midrule
		    Recall@$k$ & 1 & 2 & 4 & 8 & 1 & 2 & 4 & 8 & 1 & 2 & 4 & 8 & 1 & 2 & 4 & 8 & 1 & 2 & 4 & 8 \\
		    \midrule
		    Sum & $94.6$ & $97.2$ & $98.5$ & ${\bf 99.3}$ & $62.6$ & $71.2$ & $78.9$ & $85.0$ & $98.3$ & $99.2$ & ${\bf 99.6}$ & ${\bf 99.8}$ & $54.0$ & $68.9$ & $81.9$ & $90.6$ & $27.8$ & $37.4$ & $48.4$ & $60.4$ \\
		    Concate & ${\bf 95.1}$ & ${\bf 97.3}$ & ${\bf 98.6}$ & $99.2$ & $60.7$ & $69.8$ & $77.4$ & $83.6$ & ${\bf 98.7}$ & ${\bf 99.3}$ & ${\bf 99.6}$ & ${\bf 99.8}$ & $49.7$ & $65.3$ & $79.4$ & $88.9$ & $25.4$ & $34.6$ & $45.3$ & $57.2$ \\
		    MDL & $89.5$ & $94.1$ & $96.6$ & $98.3$ & $63.6$ & $72.6$ & $79.9$ & $86.0$ & $97.6$ & $98.8$ & $99.2$ & $99.6$ & $58.9$ & $72.9$ & $84.1$ & $92.2$ & $31.6$ & $42.0$ & $53.4$ & $64.8$ \\
		    Simple CNAPS~\cite{bateni2020improved} & $79.9$ & $86.9$ & $92.6$ & $96.2$ & $65.2$ & $73.8$ & $81.1$ & $86.6$ & $97.5$ & $98.8$ & $99.3$ & $99.7$ & ${\bf 66.2}$ & ${\bf 79.3}$ & ${\bf 88.5}$ & ${\bf 94.7}$ & $33.2$ & $44.2$ & $57.3$ & $68.7$ \\
		    Ours & $87.9$ & $93.0$ & $96.1$ & $98.2$ & ${\bf 67.4}$ & ${\bf 76.3}$ & ${\bf 83.0}$ & ${\bf 88.5}$ & $97.0$ & $98.4$ & $99.1$ & $99.5$ & $62.1$ & $76.5$ & $86.0$ & $93.3$ & ${\bf 35.1}$ & ${\bf 46.1}$ & ${\bf 57.8}$ & ${\bf 69.0}$ \\
			\bottomrule
		\end{tabular}%
			}
		\vspace{-0.25cm}
		\caption{Global retrieval performance on Meta-Dataset (unseen datasets). In addition to few-shot learning experiments, we evaluate our method in a non-episodic retrieval task to further compare the generalization ability of our universal representations. }
		\label{tab:recallunseen}
\end{table*}

\subsection{Effect of adaptors in knowledge distillation}
In this section, we evaluate our method with adaptors or without adaptors for aligning features when we use CKA for knowledge distillation. From \cref{apptab:lossf}, We can see that using adaptors can improve the performance, such as Birds (+1.7) and VGG Flower (+3.6), MSCOCO (+1.3). This indicates that the adaptors $A_{\theta}$ help align features between multi-domain and single-domain learning networks which are learned from very different domains.

\subsection{Complete results of varying-way five-shot and five-way one-shot}

We further analyze our method for 5-shot setting with varying number of categories.
To this end, we follow the setting in \cite{doersch2020crosstransformers}, compare our method to the best three state-of-the-art methods including Simple CNAPS, SUR and URT. In this setting, we sample a varying number of ways in Meta-Dataset the same as the standard setting but a fixed number of shots to form balanced support and query sets. The mean accuracy and 95\% confidence interval of our method and compared approaches are depicted in \cref{tabapp:fixedshot}. 
As shown in Table~\ref{tabapp:fixedshot}, overall performance for all methods decreases in most datasets compared to results in the conventional setting shown in Table 1 in the paper, indicating that this is a more challenging setting. It is due to that five-shot setting samples much less support images than the standard setting.
While both Simple CNAPS and SUR obtain 3.0 average rank. SUR performs the best on MNIST, Simple CNAPS outperforms others on CIFAR-100 and URT is top-1 on Quick Draw. 
Ours still achieves significant better performance than other methods on the rest ten datasets.

\paragraph{Results in five-way one-shot setting.}
Next we test an extremely challenging five-way one-shot setting on Meta-Dataset.
For each task, only one image per class is seen as support set.
This setting is often used in evaluating different methods in a single domain~\cite{Lake1332,ren2018meta,vinyals2016matching}, while we adopt it for multiple domains.
As shown in Table~\ref{tabapp:fixedshot}, our method achieves consistent gain as observed in previous two settings, which validates the importance of good universal representations in case of limited labeled samples in meta-test. Interestingly, Simple CNAPS achieves better rank than SUR in this setting, which is opposite in previous settings.

\subsection{Qualitatively results}
\begin{figure*}[ht]
\begin{center}
\includegraphics[width=0.76\linewidth]{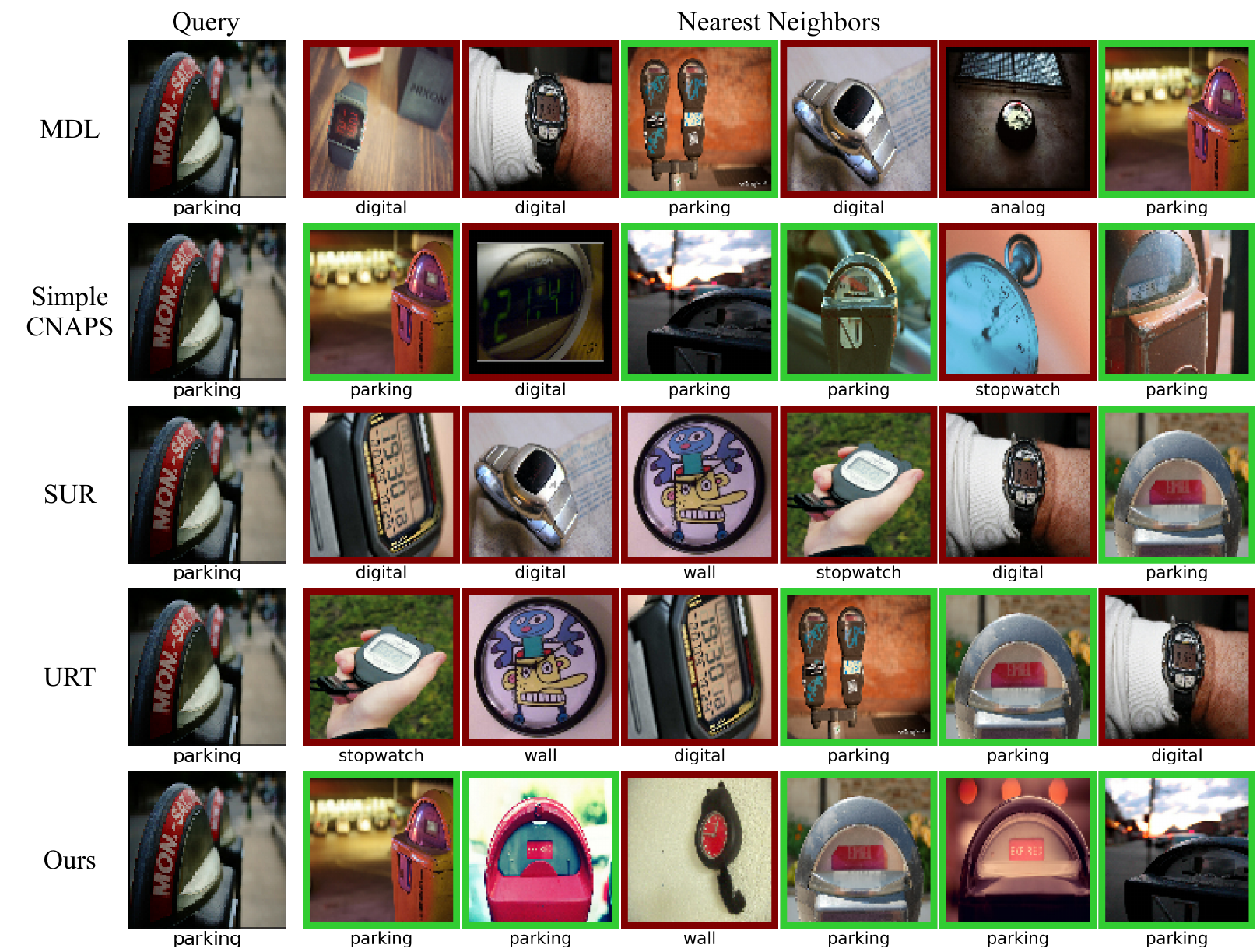}
\end{center}
\vspace{-0.45cm}
\caption{Qualitative comparison to MDL, Simple CNAPS~\cite{bateni2020improved}, SUR~\cite{dvornik2020selecting}, and URT~\cite{liu2020universal} in ImageNet. Green and red colors indicate correct and false predictions respectively.}
\label{fig:imagenet}
\end{figure*}

\begin{figure*}[ht]
\begin{center}
\includegraphics[width=0.76\linewidth]{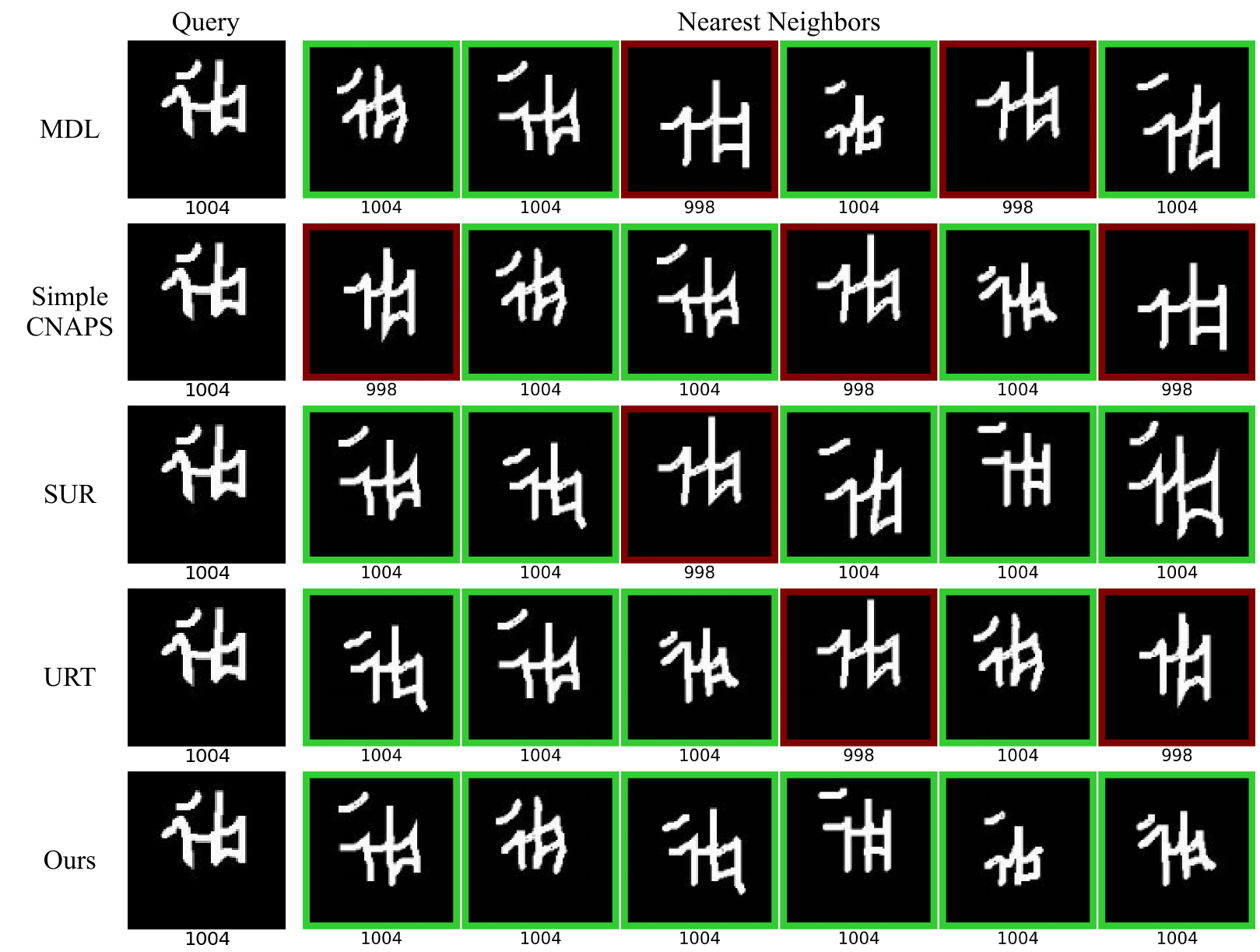}
\end{center}
\vspace{-0.45cm}
\caption{Qualitative comparison to MDL, Simple CNAPS~\cite{bateni2020improved}, SUR~\cite{dvornik2020selecting}, and URT~\cite{liu2020universal} in Omniglot. Green and red colors indicate correct and false predictions respectively.}
\label{fig:omniglot}
\end{figure*}

\begin{figure*}[ht]
\begin{center}
\includegraphics[width=0.76\linewidth]{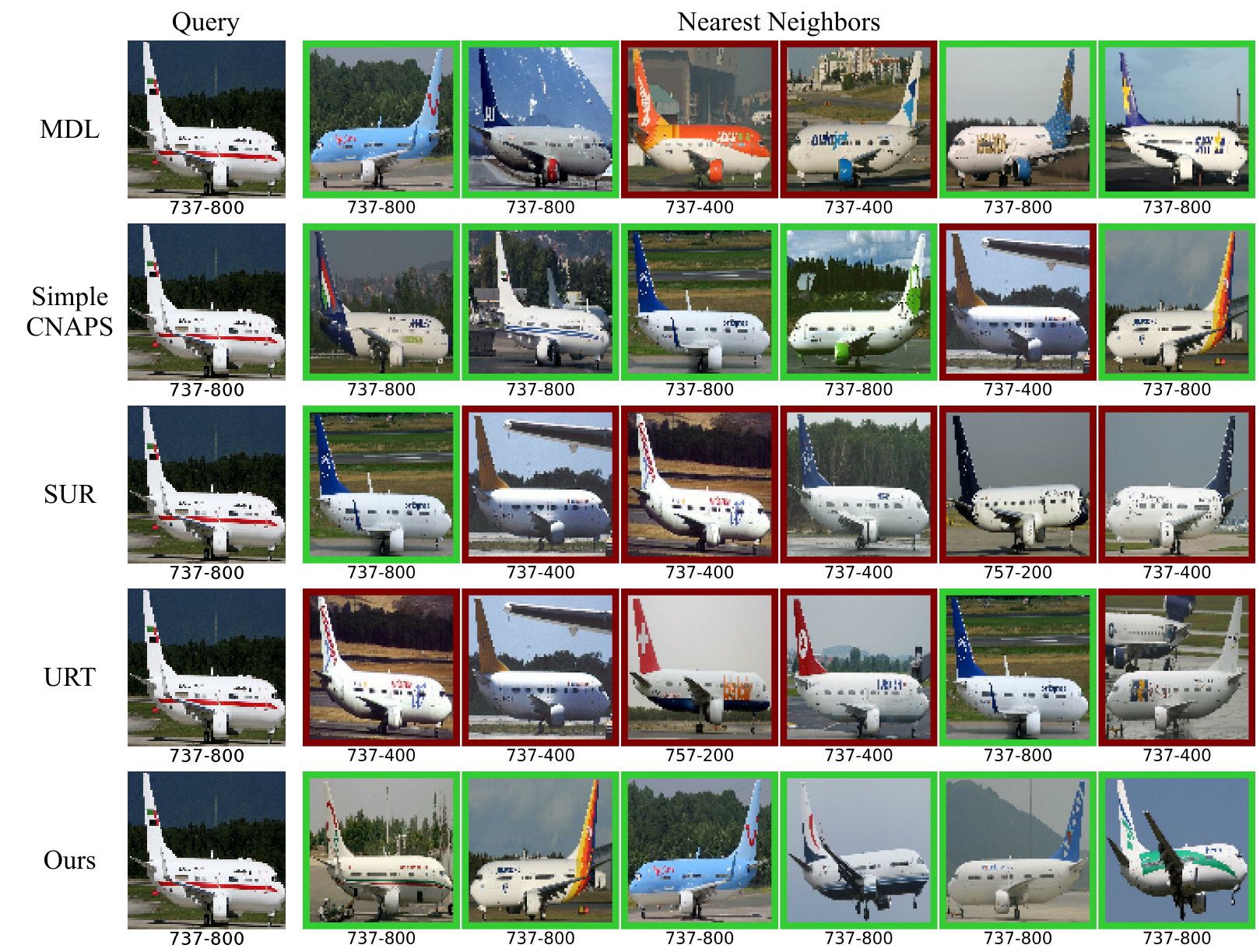}
\end{center}
\vspace{-0.45cm}
\caption{Qualitative comparison to MDL, Simple CNAPS~\cite{bateni2020improved}, SUR~\cite{dvornik2020selecting}, and URT~\cite{liu2020universal} in Aircraft. Green and red colors indicate correct and false predictions respectively.}
\label{fig:aircraft}
\end{figure*}

\begin{figure*}[ht]
\begin{center}
\includegraphics[width=0.76\linewidth]{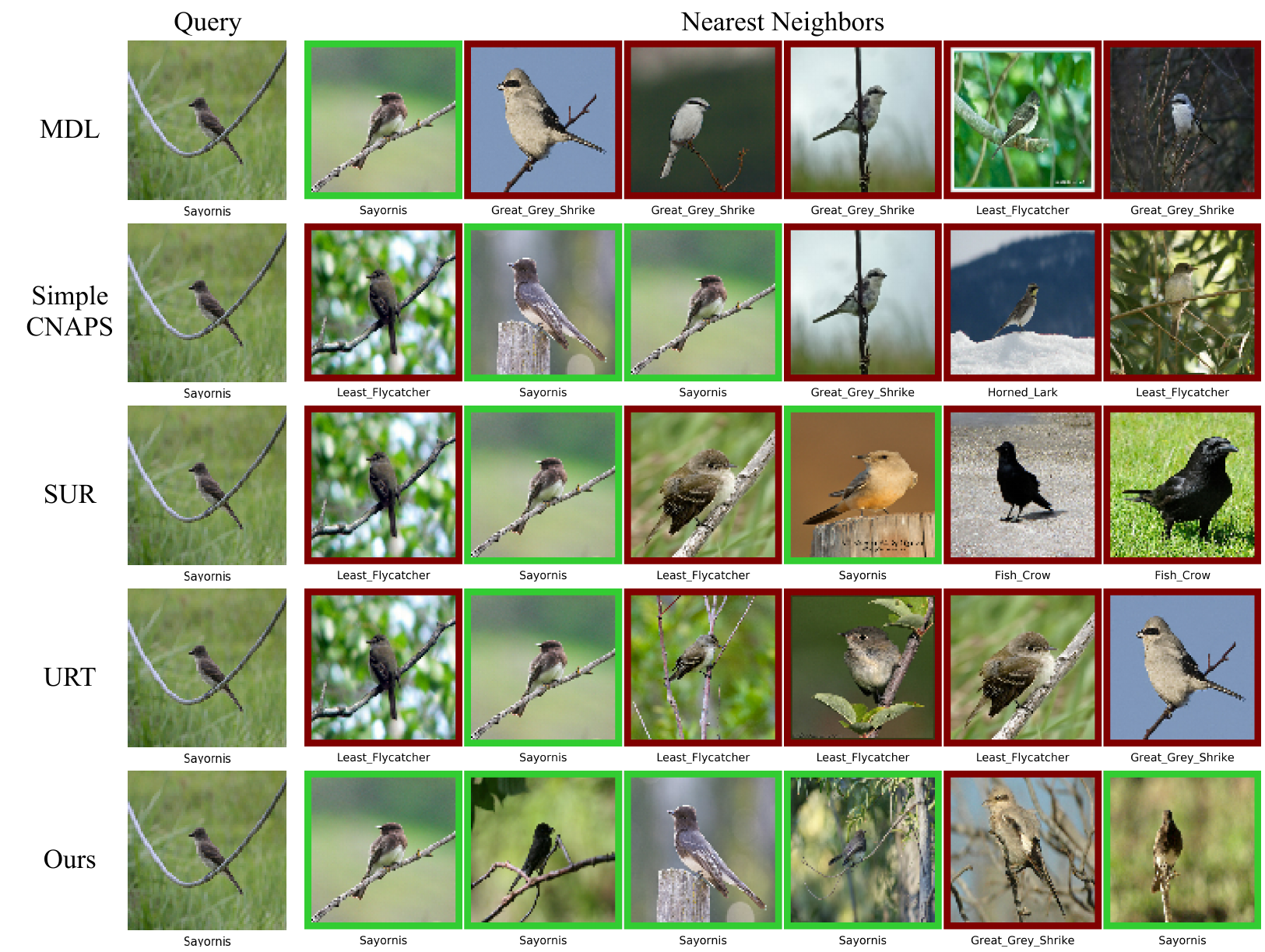}
\end{center}
\vspace{-0.45cm}
\caption{Qualitative comparison to MDL, Simple CNAPS~\cite{bateni2020improved}, SUR~\cite{dvornik2020selecting}, and URT~\cite{liu2020universal} in Birds. Green and red colors indicate correct and false predictions respectively.}
\label{fig:birds}
\end{figure*}

\begin{figure*}[ht]
\begin{center}
\includegraphics[width=0.76\linewidth]{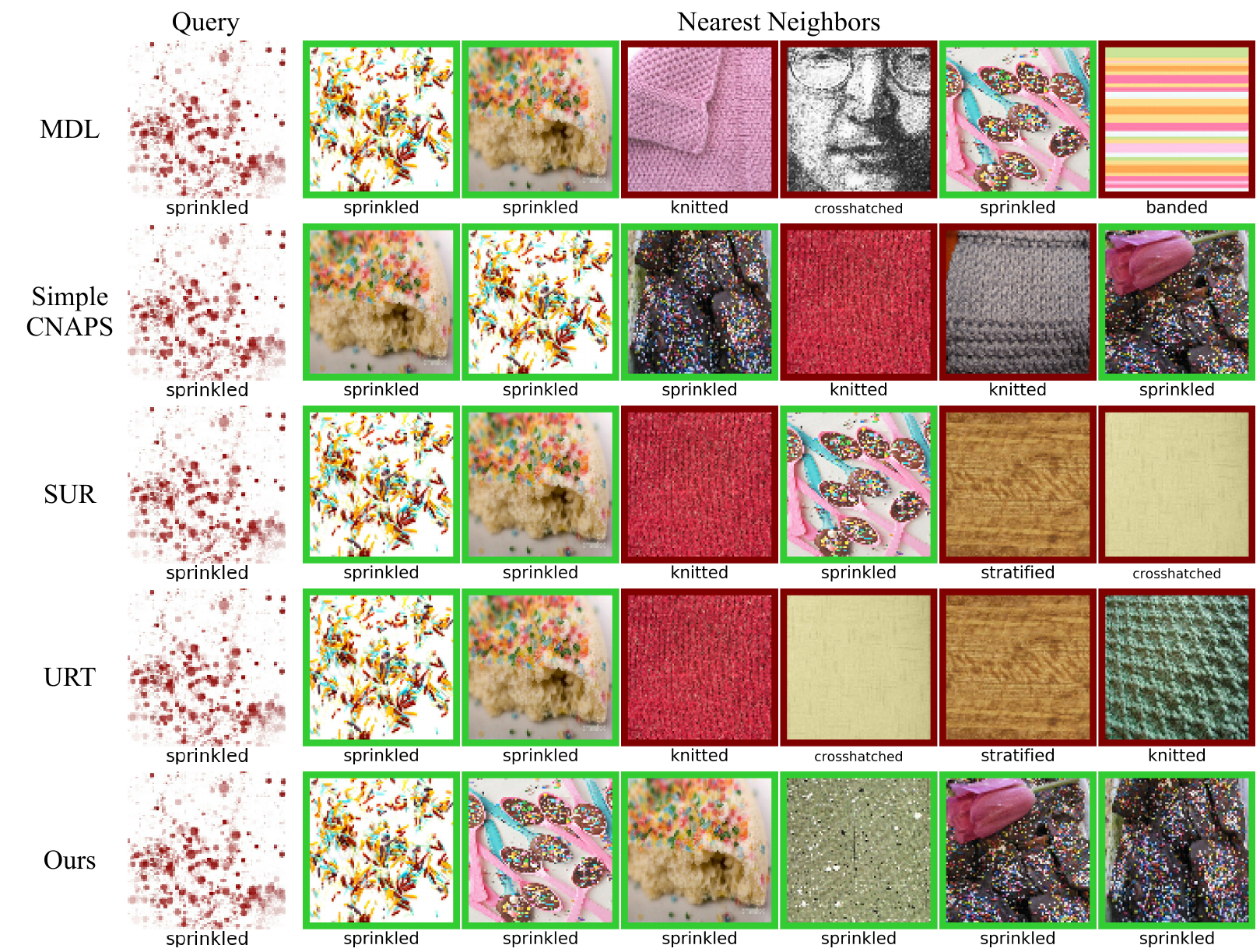}
\end{center}
\vspace{-0.45cm}
\caption{Qualitative comparison to MDL, Simple CNAPS~\cite{bateni2020improved}, SUR~\cite{dvornik2020selecting}, and URT~\cite{liu2020universal} in Textures. Green and red colors indicate correct and false predictions respectively.}
\label{fig:texture}
\end{figure*}

\begin{figure*}[ht]
\begin{center}
\includegraphics[width=0.76\linewidth]{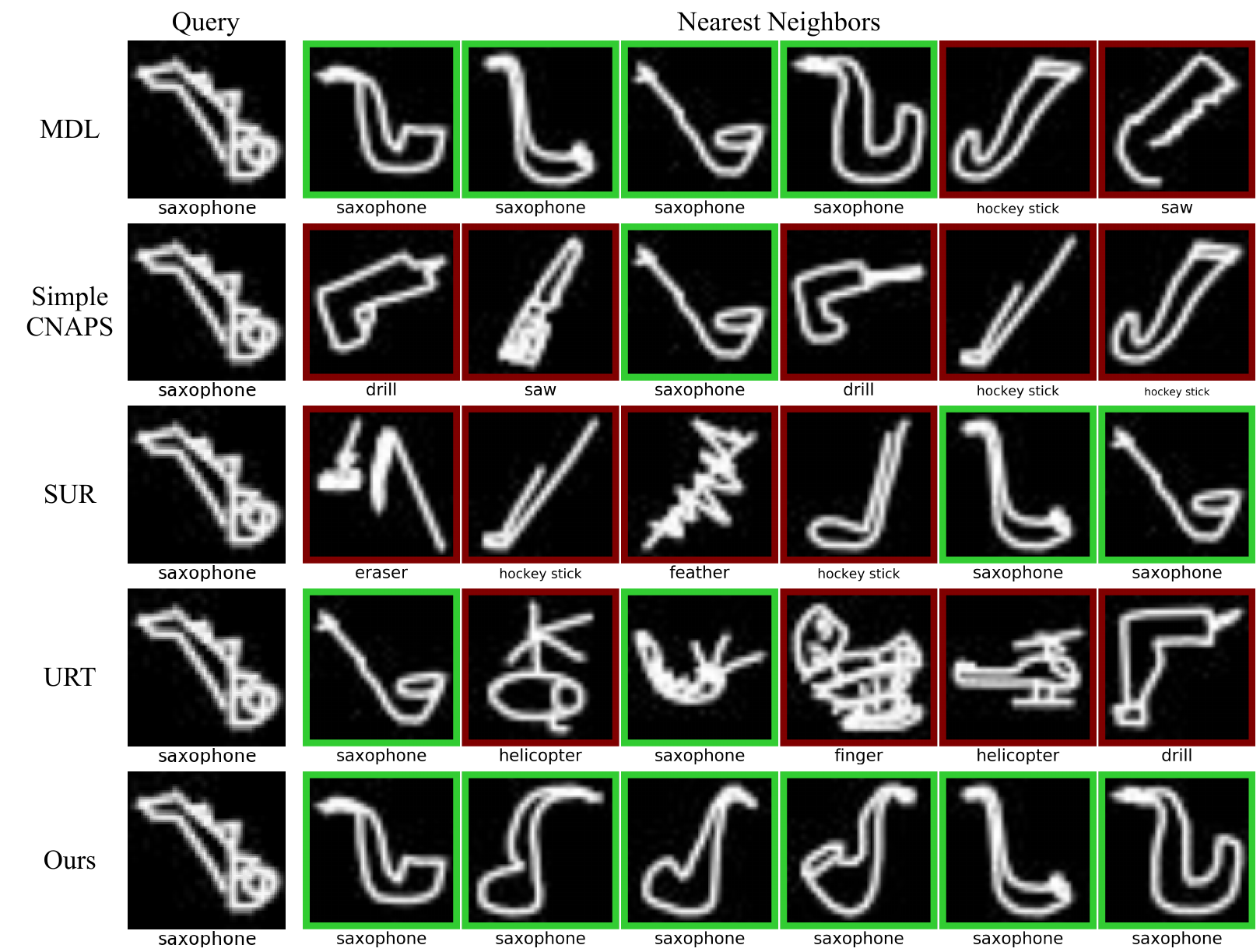}
\end{center}
\vspace{-0.45cm}
\caption{Qualitative comparison to MDL, Simple CNAPS~\cite{bateni2020improved}, SUR~\cite{dvornik2020selecting}, and URT~\cite{liu2020universal} in Quick Draw. Green and red colors indicate correct and false predictions respectively.}
\label{fig:quickdraw}
\end{figure*}

\begin{figure*}[ht]
\begin{center}
\includegraphics[width=0.76\linewidth]{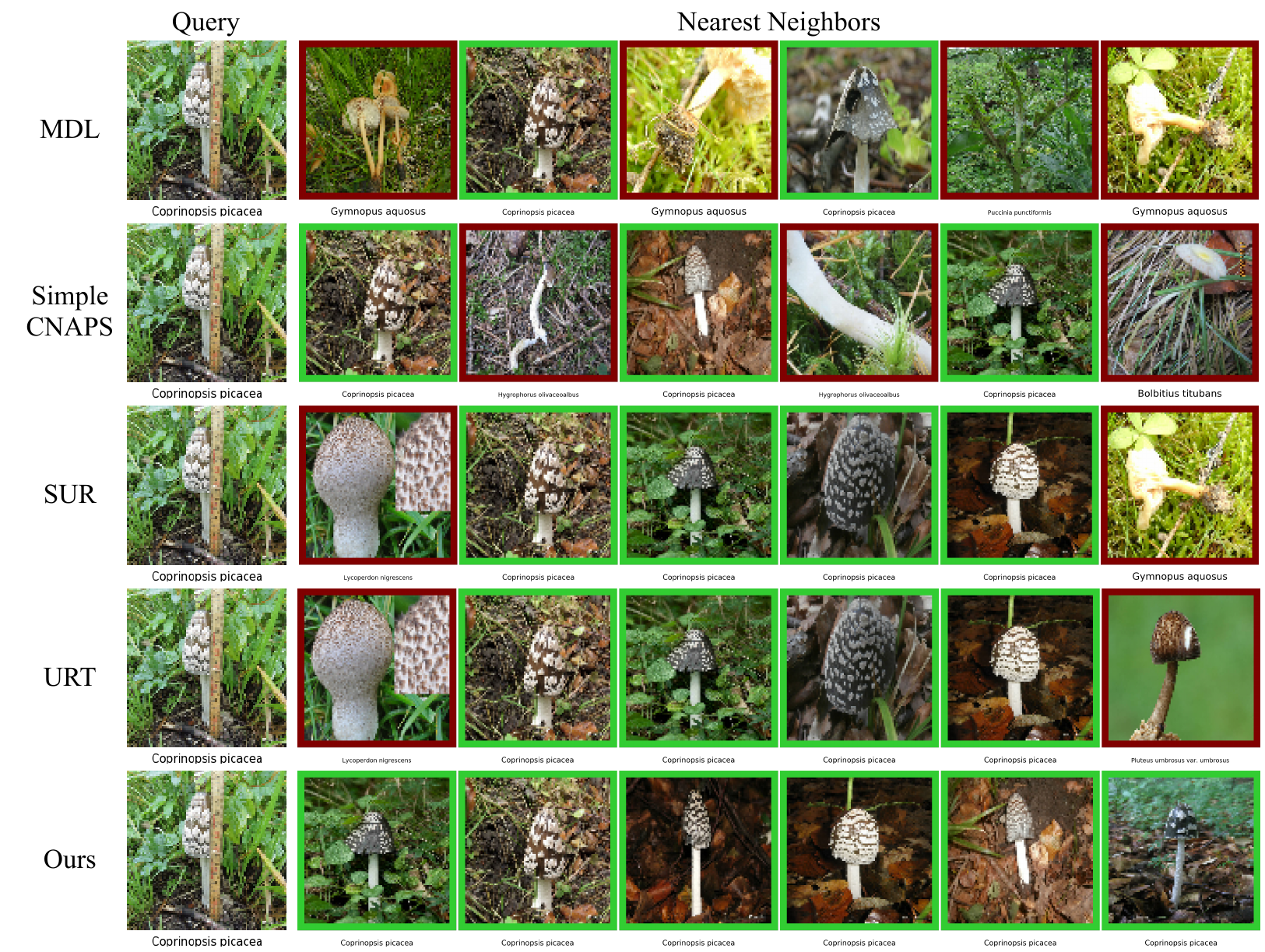}
\end{center}
\vspace{-0.45cm}
\caption{Qualitative comparison to MDL, Simple CNAPS~\cite{bateni2020improved}, SUR~\cite{dvornik2020selecting}, and URT~\cite{liu2020universal} in Fungi. Green and red colors indicate correct and false predictions respectively.}
\label{fig:fungi}
\end{figure*}

\begin{figure*}[ht]
\begin{center}
\includegraphics[width=0.76\linewidth]{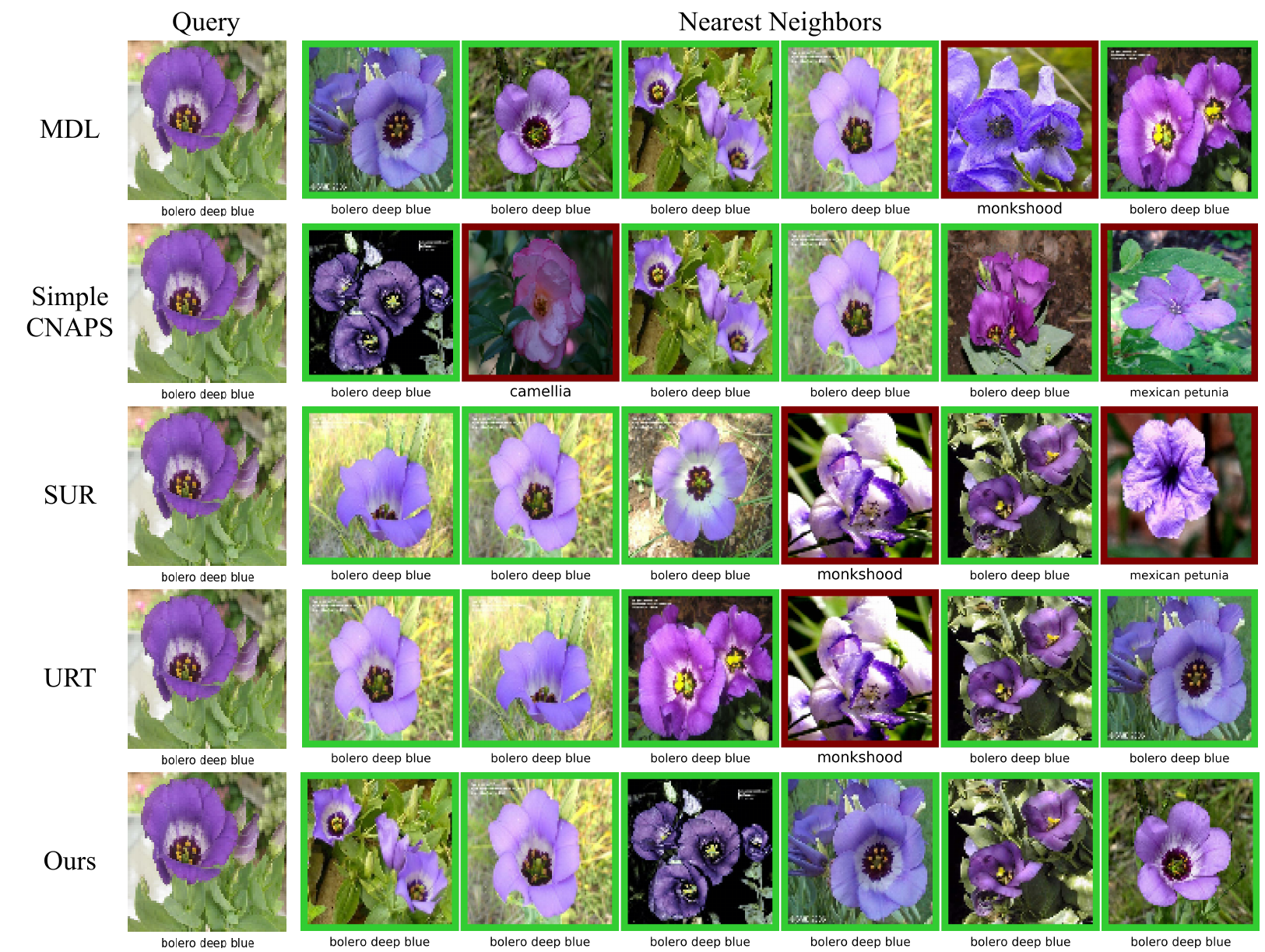}
\end{center}
\vspace{-0.45cm}
\caption{Qualitative comparison to MDL, Simple CNAPS~\cite{bateni2020improved}, SUR~\cite{dvornik2020selecting}, and URT~\cite{liu2020universal} in VGG Flower. Green and red colors indicate correct and false predictions respectively.}
\label{fig:flower}
\end{figure*}

\begin{figure*}[ht]
\begin{center}
\includegraphics[width=0.76\linewidth]{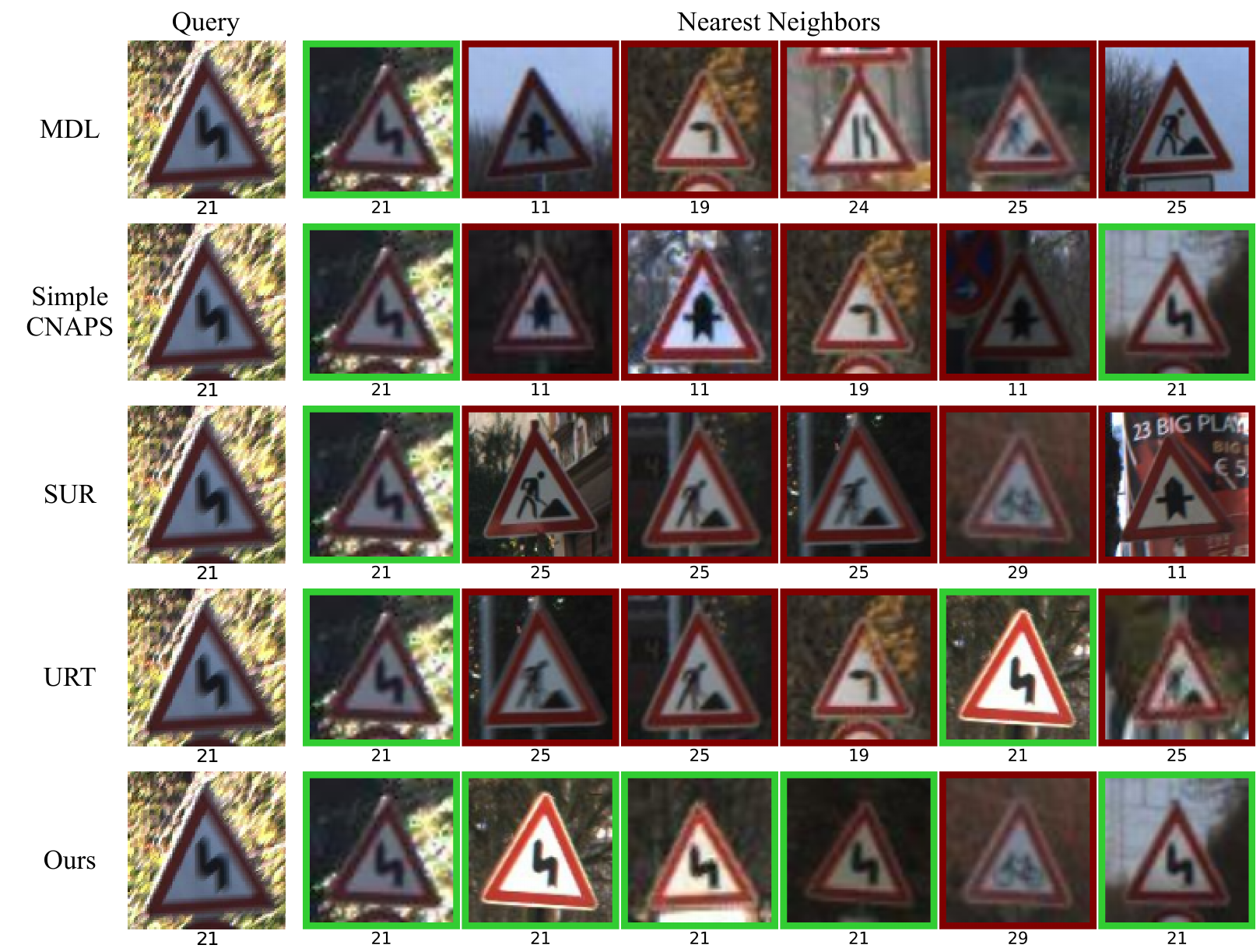}
\end{center}
\vspace{-0.45cm}
\caption{Qualitative comparison to MDL, Simple CNAPS~\cite{bateni2020improved}, SUR~\cite{dvornik2020selecting}, and URT~\cite{liu2020universal} in Traffic Sign. Green and red colors indicate correct and false predictions respectively.}
\label{fig:traffic}
\end{figure*}

\begin{figure*}[ht]
\begin{center}
\includegraphics[width=0.76\linewidth]{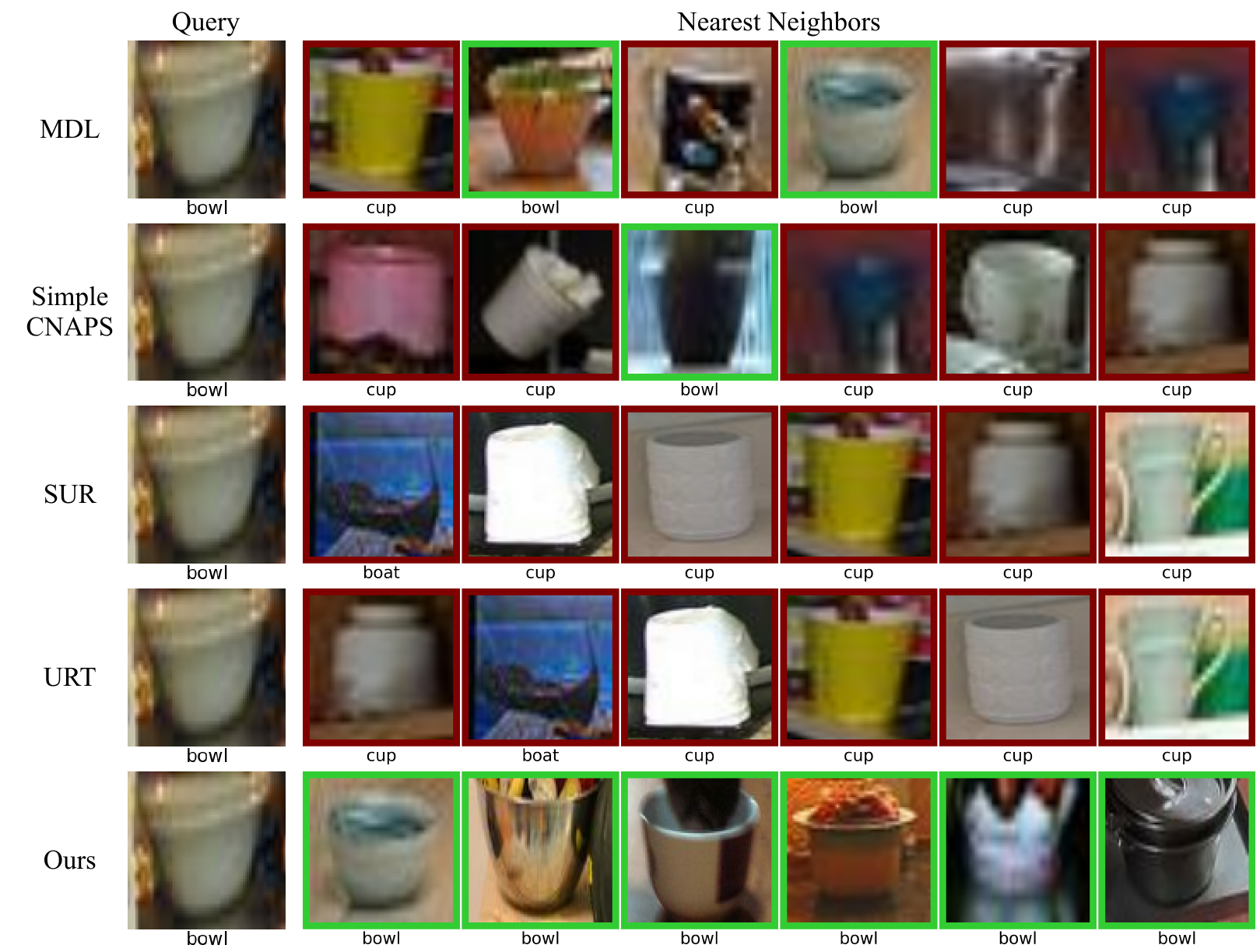}
\end{center}
\vspace{-0.45cm}
\caption{Qualitative comparison to MDL, Simple CNAPS~\cite{bateni2020improved}, SUR~\cite{dvornik2020selecting}, and URT~\cite{liu2020universal} in MSCOCO. Green and red colors indicate correct and false predictions respectively.}
\label{fig:mscoco}
\end{figure*}

\begin{figure*}[ht]
\begin{center}
\includegraphics[width=0.76\linewidth]{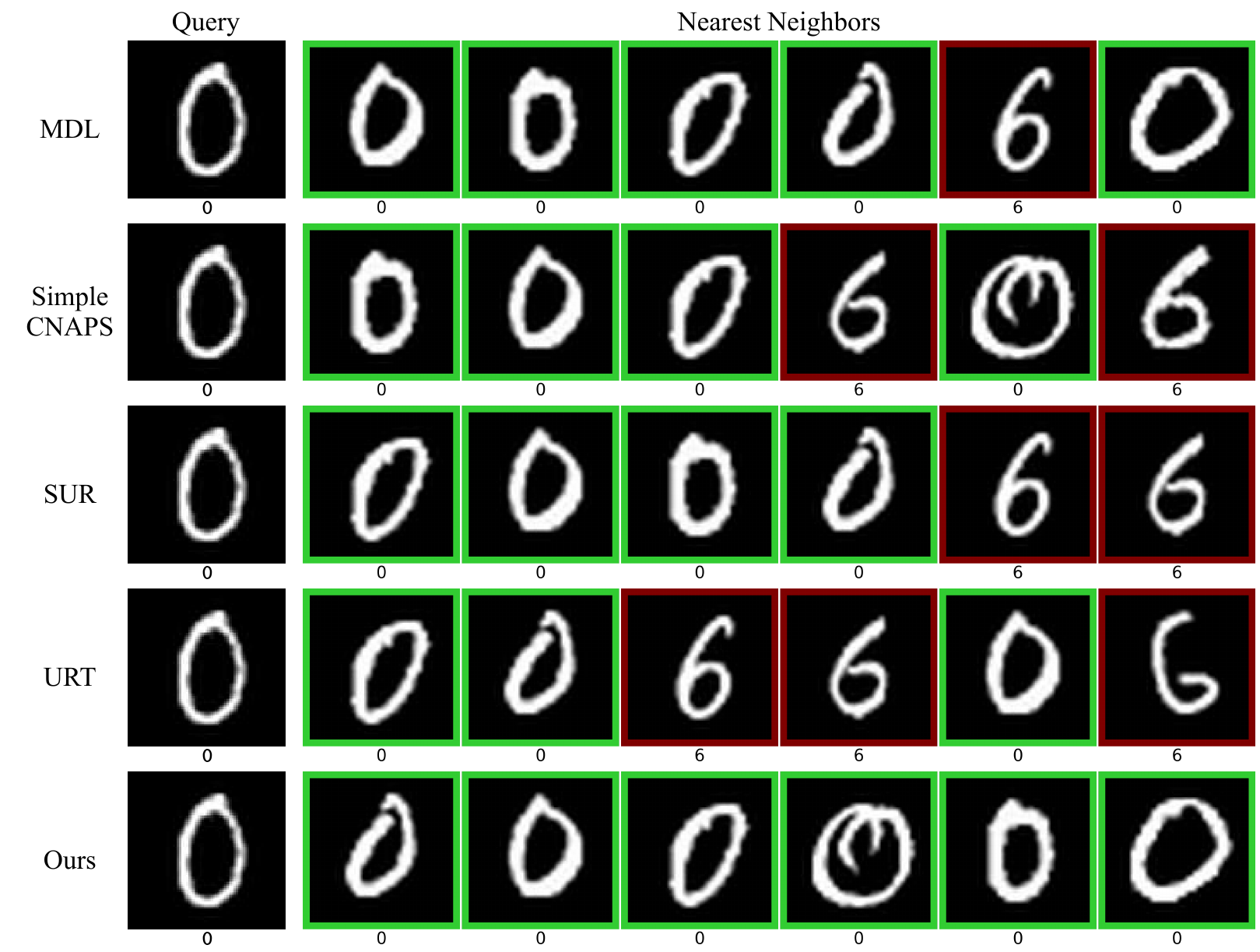}
\end{center}
\vspace{-0.45cm}
\caption{Qualitative comparison to MDL, Simple CNAPS~\cite{bateni2020improved}, SUR~\cite{dvornik2020selecting}, and URT~\cite{liu2020universal} in MNIST. Green and red colors indicate correct and false predictions respectively.}
\label{fig:mnist}
\end{figure*}

\begin{figure*}[ht]
\begin{center}
\includegraphics[width=0.76\linewidth]{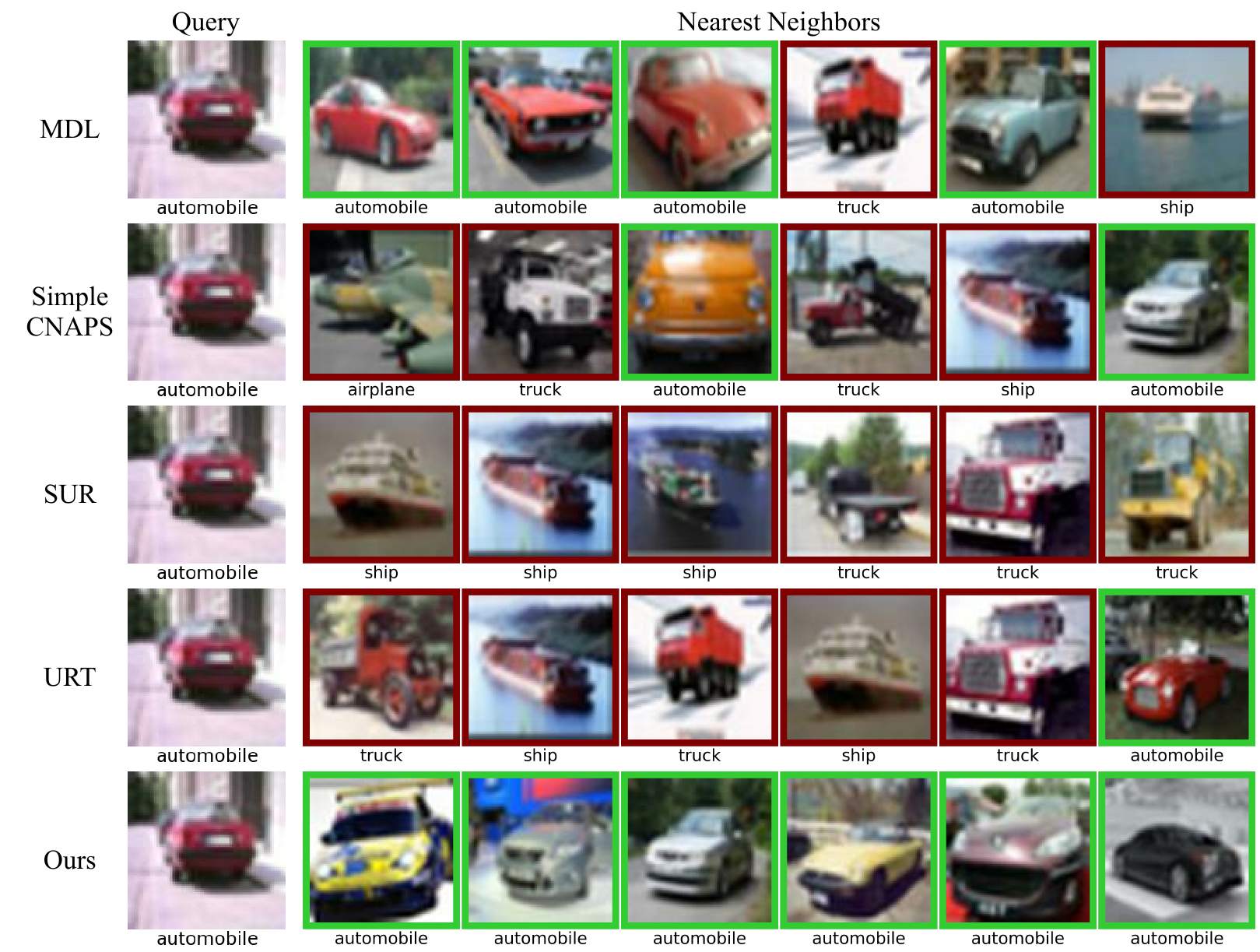}
\end{center}
\vspace{-0.45cm}
\caption{Qualitative comparison to MDL, Simple CNAPS~\cite{bateni2020improved}, SUR~\cite{dvornik2020selecting}, and URT~\cite{liu2020universal} in CIFAR-10. Green and red colors indicate correct and false predictions respectively.}
\label{fig:cifar10}
\end{figure*}

\begin{figure*}[ht]
\begin{center}
\includegraphics[width=0.76\linewidth]{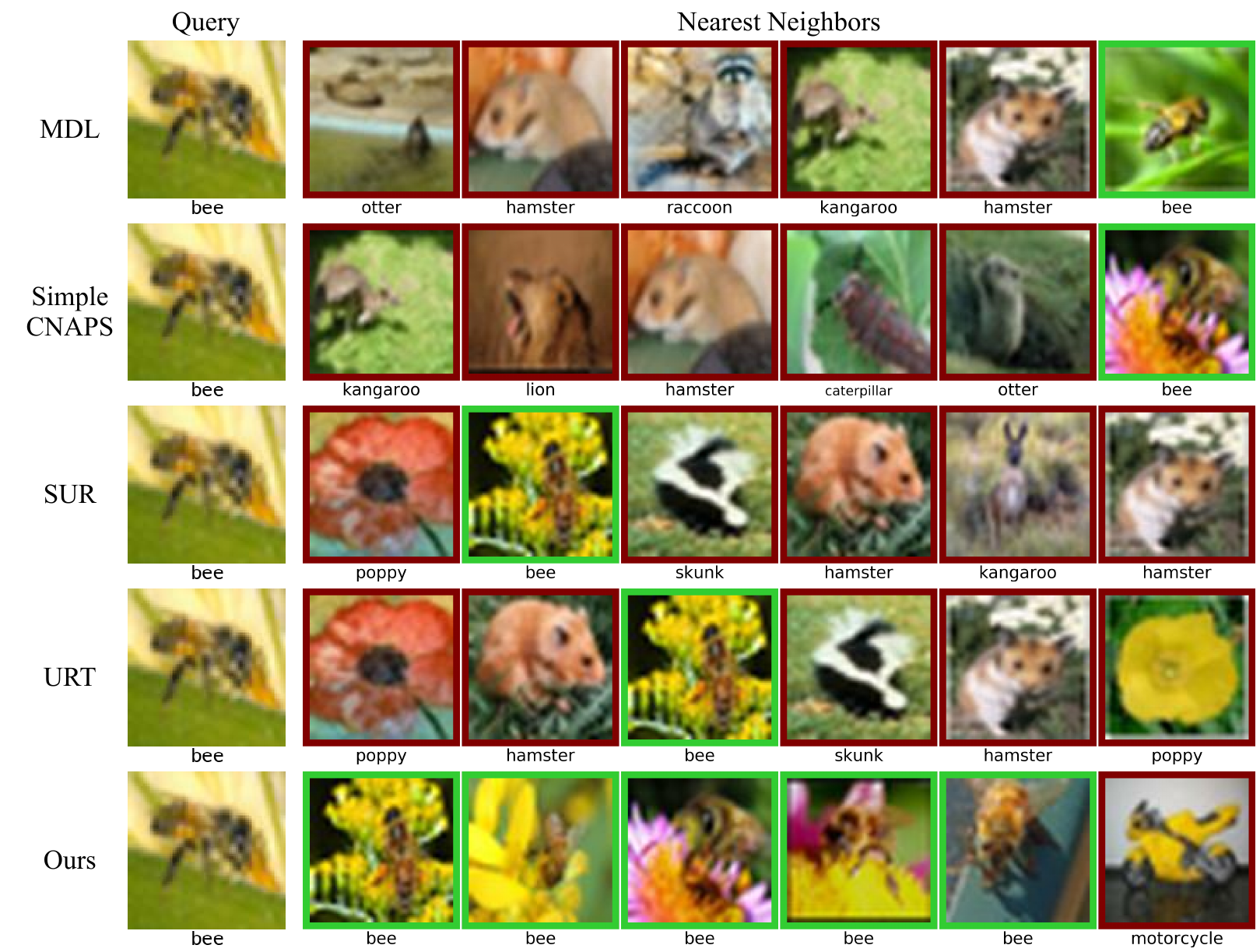}
\end{center}
\vspace{-0.45cm}
\caption{Qualitative comparison to MDL, Simple CNAPS~\cite{bateni2020improved}, SUR~\cite{dvornik2020selecting}, and URT~\cite{liu2020universal} in CIFAR-100. Green and red colors indicate correct and false predictions respectively.}
\label{fig:cifar100}
\end{figure*}

We qualitatively analyze our method and compare it to the vanilla multi-domain leanring (MDL) baseline, Simple CNAPS~\cite{bateni2020improved}, SUR~\cite{dvornik2020selecting} and URT~\cite{liu2020universal} in \cref{fig:imagenet,fig:omniglot,fig:aircraft,fig:birds,fig:texture,fig:quickdraw,fig:fungi,fig:flower,fig:traffic,fig:mscoco,fig:mnist,fig:cifar10,fig:cifar100} by illustrating the nearest neighbors in all test datasets given a query image.
It is clear that our method produces more correct neighbors than other methods. 
While other methods retrieves images with more similar colors, shapes and backgrounds, \eg in \cref{fig:traffic,fig:mscoco}, our method is able to retrieve semantically similar images. It again suggests that our method is able to learn more useful and general representations.

\subsection{Complete global retrieval results}
Here we go beyond the few-shot classification experiments and evaluate the generalization ability of our representations that are learned in the multi-domain network in a retrieval task, inspired from metric learning literature~\cite{oh2016deep,yu2019learning}.
To this end, for each test image, we find the nearest  images in entire test set in the feature space and test whether they correspond to the same category.
For evaluation metric, we use Recall@$k$ which considers the predictions with one of the $k$ closest neighbors with the same label as positive.
In \cref{tab:recallseen,tab:recallunseen}, we compare our method with Simple CNAPS in Recall@1, Recall@2, Recall@4 and Recall@8. URT and SUR require adaption using support set and no such adaptation in retrieval task is possible, we replace them with two baselines that concatenate or sum features from multiple domain-specific networks. 
Our method achieves the best performance in ten out of thirteen domains with significant gains in Aircraft, Birds, Textures and Fungi.
This strongly suggests that our multi-domain representations are the key to the success of our method in the previous few-shot classification tasks.

\end{document}